\documentclass{article}
\usepackage{ijcai20}

\usepackage{times}
\usepackage{soul}
\usepackage{url}
\usepackage{hyperref}
\usepackage[utf8]{inputenc}
\usepackage[small]{caption}
\usepackage{graphicx}
\usepackage{amsmath}
\usepackage{amsthm}
\usepackage{booktabs}
\usepackage{latexsym}
\usepackage{xspace}

\usepackage{graphicx}
\usepackage{xcolor}

\usepackage{mathtools}
\usepackage{amssymb}
\usepackage{mysymbols}
\usepackage{comment}
\usepackage{makecell}
\usepackage{paralist}
\usepackage{enumitem}
\usepackage{subcaption}

\usepackage[ruled,vlined,oldcommands,linesnumbered]{algorithm2e}

\belowdisplayskip \abovedisplayskip

\usepackage{titlesec}
\titlespacing*{\section}{0pt}{1ex plus 0ex minus 0.5ex}{0.5ex plus 0ex minus 0.5ex}
\titlespacing*{\subsection}{0pt}{1ex plus 0ex minus 0.5ex}{0.5ex plus 0ex minus 0.5ex}
\newlength{\sectionReduceTop}
\newlength{\sectionReduceBot}
\newlength{\subsectionReduceTop}
\newlength{\subsectionReduceBot}
\newlength{\abstractReduceTop}
\newlength{\abstractReduceBot}
\newlength{\captionReduceTop}
\newlength{\captionReduceBot}
\newlength{\subsubsectionReduceTop}
\newlength{\subsubsectionReduceBot}

\newlength{\eqnReduceTop}
\newlength{\eqnReduceBot}

\newlength{\horSkip}
\newlength{\verSkip}

\newlength{\figureHeight}
\makeatletter
\renewcommand{\paragraph}{%
  \@startsection{paragraph}{4}%
  {\z@}{0.75ex \@plus 0ex \@minus 0ex}{-1em}%
  {\normalfont\normalsize\bfseries}%
}
\makeatother
\newcommand{\ourdata}{novel pair\xspace}
\newcommand{\emnlpdata}{novel instance\xspace}
\newcommand{\qbot}{Q-bot\xspace}
\newcommand{\abot}{A-bot\xspace}
\newcommand{\xhdr}[1]{\vspace{3pt}\noindent\textbf{#1}}

\let\citep\cite
\let\citet\cite

\title{Emergence of Compositional Language with Deep Generational Transmission}

\author{
Michael Cogswell$^1$
\and
Jiasen Lu$^1$\and
Stefan Lee$^2$\and
Devi Parikh$^{1,3}$\and
Dhruv Batra$^{1,3}$
\affiliations
$^1$Georgia Tech
$^2$Oregon State University
$^3$Facebook AI Research
\emails
\{cogswell, jiasenlu, parikh, dbatra\}@gatech.edu,
leestef@oregonstate.edu
}

\begin{document}

\maketitle

\begin{abstract}
\vspace{-5pt}
Recent work has studied the emergence of language among deep reinforcement learning agents that must collaborate to solve a task. Of particular interest are the factors that cause language to be compositional---\ie express meaning by combining words which themselves have meaning. Evolutionary linguists have found that in addition to structural priors like those already studied in deep learning, the dynamics of transmitting language from generation to generation contribute significantly to the emergence of  compositionality. In this paper, we introduce these cultural evolutionary dynamics into language emergence by periodically replacing agents in a population to create a knowledge gap, implicitly inducing cultural transmission of language. We show that this implicit cultural transmission encourages the resulting languages to exhibit better compositional generalization.
\vspace{-4pt}
\end{abstract}

\section{Introduction} \label{sec:intro}

Compositionality is an important structure of language that reflects a disentangled understanding of the world
-- enabling the expression of infinitely many concepts 
using finitely many elements. Agents that have compositional
understandings of the world generalize in obviously 
correct ways even in the face of limited training examples~\citep{lake18generalization}. 
For example, an agent with a compositional understanding of \texttt{blue squares} 
and \texttt{purple triangles} should also understand \texttt{purple squares} 
without directly observing any of them. Developing
artificial agents that can ground, understand, and produce compositional (and therefore more interpretable)
language could greatly improve generalization to new instances and ease human-AI interactions. 

In building theories of how compositionality emerges in human languages, work in evolutionary linguistics looks to the process of cultural transmission \citep{kirby01,kirby08}. Cultural transmission of language occurs when a group of
agents pass their language on to a new group of agents,
\eg parents who teach their children to speak as they do.
Because this education is incomplete and biased, it allows the
language itself to change over time via a process known as 
\emph{cultural evolution}. This paradigm~\citep{kirby_survey14}
explains the emergence of compositionality as a result of
expressivity and compressibility -- \ie to be most 
effective, a language should be expressive enough to differentiate 
between all possible meanings (\eg, objects) and compressible enough 
to be learned easily. Work in the evolutionary linguistics community 
has shown that over multiple `generations' these competing pressures result in the 
emergence of compositional languages both in simulation~\citep{kirby01} 
and with human subjects~\citep{kirby08}. These studies aim to understand humans
whereas we want to understand and design artificial neural networks.

\begin{figure}[t]
    \centering
    \vspace{-20pt}
    
    \begin{subfigure}[b]{0.35\textwidth}
        \includegraphics[width=1\linewidth,trim={20pt 30pt 10pt 20pt},clip]{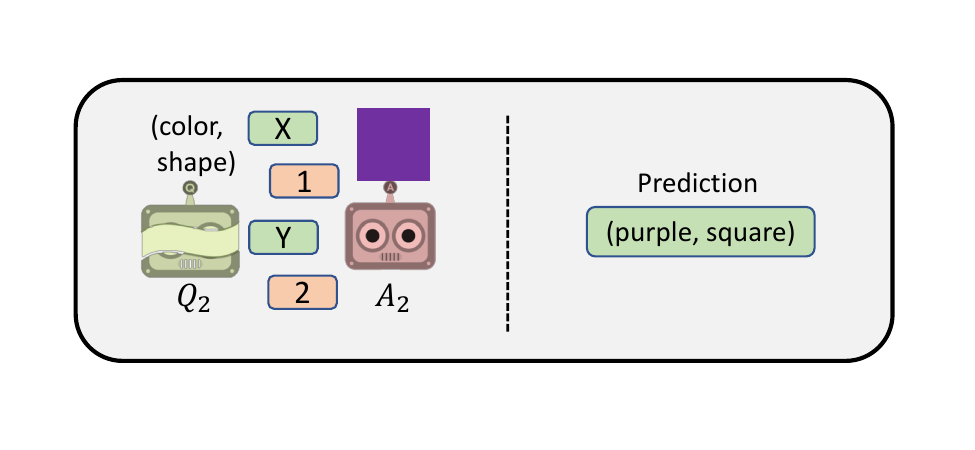}
    \end{subfigure}%
    
    \begin{subfigure}[b]{0.48\textwidth}
        \includegraphics[width=1\linewidth,trim={20pt 20pt 0pt 40pt},clip]{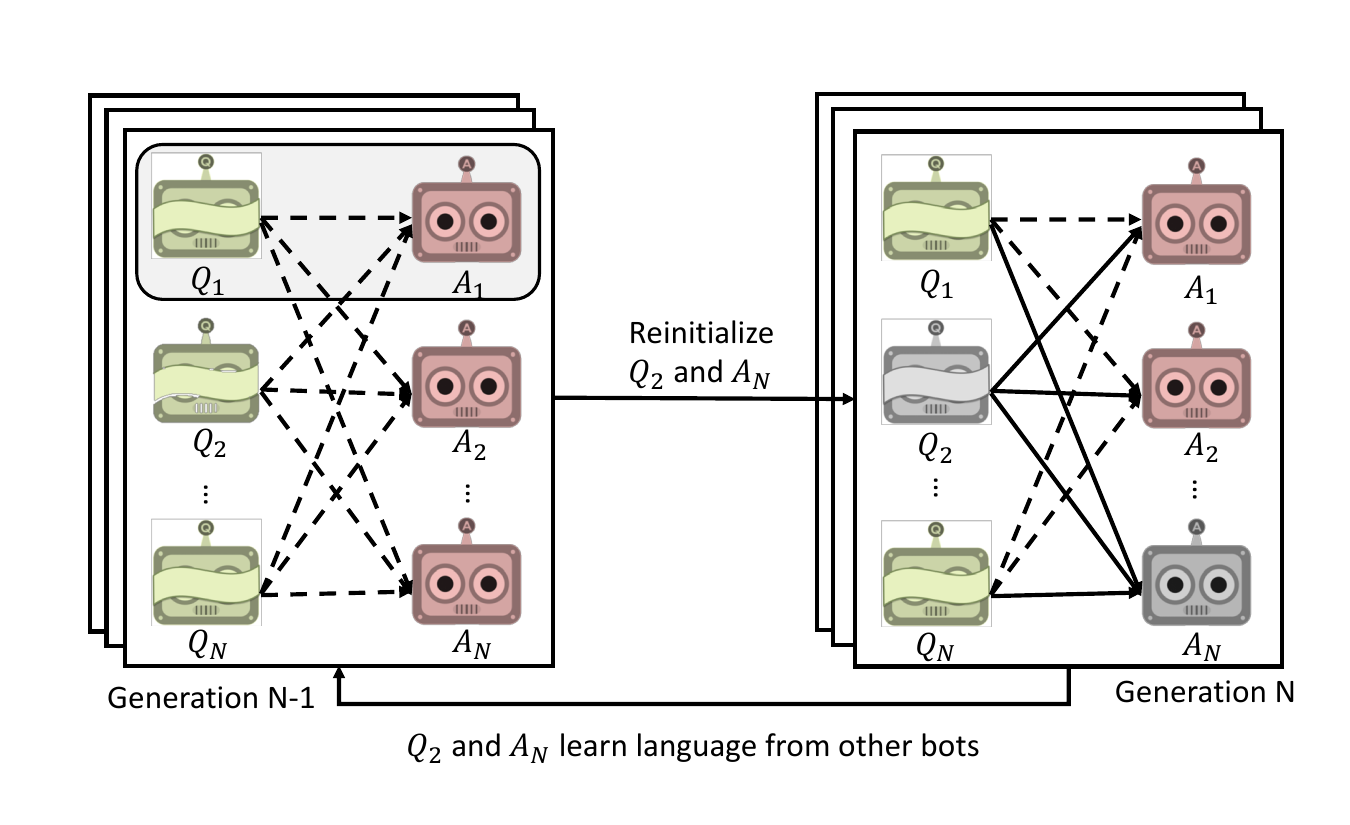}
    \end{subfigure}
    
    \caption{We introduce cultural transmission into language
    emergence between neural agents. 
    Start with the goal-oriented dialog task at the top of the figure (similar to that of \protect\citet{emnlp}).
    During learning we periodically
    replace some agents with new ones (gray agents). These new agents
    do not know any language, but instead of creating one
    they learn it from older agents.
    This creates generations of language that become more
    compositional over time.}
    \label{fig:teaser}
\end{figure}

Approaching the problem from another direction, recent work in AI has studied language emergence 
in such multi-agent, goal-driven tasks. These works have demonstrated that agent languages
will emerge to enable coordination-centric tasks to be solved without 
direct or even indirect language supervision~\citep{foerster16,fergus16,baroni16,visdial_rl}.
However, the resulting languages are usually not compositional and are  
difficult to interpret, even by other machines \citep{translating_neuralese}.
Some existing work has studied means to encourage compositional language formation
\citep{mordatch_abbeel,emnlp}, but these settings study fixed populations of agents  -- \ie examining language within a single generation.

\textbf{In this work we bridge these two areas -- examining the effect of 
generational cultural transmission on the compositionality of emergent languages
in a multi-agent, goal-driven setting. }

We study this in the context of a cooperative dialog-based reference game involving two agents communicating in discrete symbols \citep{emnlp}; an example dialog is shown at the top of \figref{fig:teaser}. To examine cultural transmission, we extend this setting to a population of agents (bottom of \figref{fig:teaser}) and introduce a simple mechanism to induce the expressivity and compressibility pressures inherent in cultural transmission. Specifically, we periodically re-initialize some subset of the agents in the population. In order to perform well at the task, the population's emergent language must be sufficiently expressive to reference all the objects (expressivity) and must be easily learnable by these `new' agents (compressibility).
The new agents have a randomized language whereas the surviving agents already know a grounded language. This ``knowledge gap'' creates an implicit `teaching' setting that is analogous to the explicit transmission stage in models of \emph{iterative learning} \citep{kirby01}.


Through  our  experiments  and  analysis,  we  show  that  periodic  agent  replacement  is  an  effective way to induce cultural transmission and yields more compositionally generalizable language in our setting.
To summarize, our contributions are:
\begin{compactitem}[--]
    \item
    We propose a method for inducing implicit cultural transmission in neural language models.
    \item
    We measure the similarity between agent languages and verify cultural transmission has occurred as a result of our periodic agent replacement protocol.
    \item
    We show our cultural transmission procedure induces compositionality in neural language models,
    going from 13\% accuracy on a compositionally novel test set to 46\% in the best configuration.
    Further, we show this is complementary with previous priors which encourage compositionality.
\end{compactitem}




\section{\fontsize{11.5}{12}\selectfont Task \& Talk: A Testbed for Compositional Language Emergence} \label{sec:approach_dialog}

We consider the cooperative \emph{Task \& Talk} reference game 
introduced in \cite{emnlp}. Shown in the top of \figref{fig:teaser}, the game
is played by two agents -- one who observes an attributed object 
-- \eg \texttt{(purple, solid, square)} -- and another who is given a task 
to retrieve a subset of these attributes over the course of the dialog -- \eg \texttt{(color,shape)}. The dialog itself consists of two rounds
of agents exchanging single-token utterances from fixed vocabularies. At
the end of the dialog, the task-aware agent must report the requested
attributes and both agents are rewarded for correct predictions.
This causes a language grounded in the objects to \emph{emerge}
because there is no other way to solve the task.

A compositional solution to this task can look like a question-answer
style dialog where the task-aware agent queries the other for specific attributes (top of \figref{fig:teaser})
-- \eg uttering ``X'' requesting the \texttt{color} to which the other agent 
replies ``1'' indicating \texttt{purple}. Importantly, this pattern would
persist regardless of the other attribute values of the object
(\eg for all \texttt{(purple, *, *)} objects).
However, as there is no grounding supervision 
provided, agents must learn to associate specific meanings to specific words and it is unlikely for compositional languages to emerge purely by chance.
Given the same color task, an non-compositional agent might use ``1''
for \texttt{(purple, solid, square)} and then ``2'' for
a novel instance \texttt{(purple, solid, circle)}. Other agents have no way to know 
that ``2'' means purple instead of ``1'', so
compositional language is essential for generalization to compositionally novel instances.

\xhdr{Models.} To formalize this setting, let \qbot{} and \abot{} be agent policies
parameterized by neural networks $Q$ and $A$ respectively. At each round $t$, \qbot{} 
observes the task $x_Q$ and it's memory of the dialog so far $h_Q^{t-1}$ and produces
a single-token utterance $m_Q^{t}\in \mathcal{V}$ from the vocabulary $\mathcal{V}$.
Functionally, $m_Q^{t}, h_Q^t = Q(m_A^{t-1}, x_Q, h_Q^{t-1})$ where $m_A^{t-1}$ is \abot{}'s
reply in the previous round. Likewise, \abot{} responds by computing 
$m_A^{t}, h_A^t = A(m_Q^{t}, x_A, h_A^{t-1})$ where $x_A$ is the object instance represented 
symbolically by concatenating 3 one-hot vectors, one per attribute. After two rounds, \qbot{} must respond
to the task, predicting the requested attribute pair $\hat{u} = U(x_Q, h_Q^T)$ as a function
of the task and \qbot{}'s final memory state. Both agents are rewarded if both attributes are correct 
(no partial credit). We follow the neural network architectures of $Q, A$, and $U$ 
from \cite{emnlp}.

\xhdr{Measuring Compositional Generalization.} \citet{emnlp} generated a synthetic
dataset consisting of three attribute types (\texttt{color, shape, style}) each with four values (\eg, \texttt{red}, \texttt{blue}, \texttt{square}, \texttt{star}, \texttt{dotted}, \texttt{solid}, ...) and six tasks, one task for each ordered pair of different attribute types. This results in 64 unique instances and 384 task-instance pairs. To evaluate compositionality, \citet{emnlp} held out 12 random instances for testing. Given the closed-world set of instances, these 12 triplets of attributes is not seen during training; however, each individual value is seen in other triplets that do appear in training. As such, test accuracy is a measure of compositional generalization.

%
%
%
%
%
%

\xhdr{Shortcomings of \cite{emnlp} Evaluation.} In our investigations, we found some shortcomings in the evaluation protocol of \cite{emnlp}. First, the authors do not report variance over multiple runs or different random test-sets which we found to be significant. Second, the strategy of randomly selecting the test set can still reward some only partially compositional strategies.
For instance, suppose agents develop a language that uses single words to refer to attribute pairs 
like \texttt{(red, *, triangle)} and \texttt{(red, filled, *)}. Such agents might generalize to an unseen 
instance \texttt{(red, filled, triangle)} by composing the `paired' words above instead of 
disentangling individual attributes. 

We make two modifications to address these issues. Our results are reported as means and variances 
estimated from multiple training runs with 4 different random seeds and 4-way 
cross-validation (16 experiments). We also introduce a harder dataset where instead of withholding random
individual instances (e.g., \texttt{(green,dotted,triangle)},...) as in \cite{emnlp},
we withhold  all instances for a set of attribute pairs (e.g., 
\texttt{(green,dotted,*)},\texttt{(red,solid,*)},...).
We will refer to datasets generated in this fashion as \textbf{\ourdata} and the original dataset
as \textbf{\emnlpdata}. We report on both settings for comparison (see appendix A.1), but find our new setting to be
significantly more challenging in practice -- requiring a stricter notion of compositionality more closely aligned with human intuitions about these attributes.

\section{\fontsize{11.5}{12}\selectfont Compositional Language Emergence with Cultural Transmission}
\label{sec:approach_cultural_transmission}

In iterative learning models of cultural transmission from evolutionary linguistics,  competing pressures towards expressivity and compressibility have been shown to induce compositionality over multiple `generations' of language transfer \citep{kirby01,kirby08}. The goal-driven nature of our reference game already encourages expressivity -- agents must be able to refer to the objects in order to succeed. To introduce compressibility pressure and parallel literature in evolutionary linguistics, we introduce a population of agents which regularly has members replaced by new agents that lack any understanding of the remaining population's language. As this paradigm lacks explicit teaching steps where new agents are trained to ground existing words, we consider this approach as a means of  implicit cultural transmission.
\vspace{-5pt}


\begin{algorithm}
\caption{
Training with \textcolor{blue}{Replacement} and \textcolor{red}{Multiple Agents}}
\label{alg:train_aug}
\For{epoch $e=1, \ldots, N_{\text{epochs}}$}{
    \textcolor{red}{Sample \qbot{} $i_Q$ from $\calU\{1, N_Q\}$ and \abot{} $i_A$ from $\calU\{1, N_A\}$} \\
    \For{$x_Q, x_A, u$ in each batch}{
        \For{dialog rounds $t=1,\ \ldots \,T$}{
            $m_Q^t, h_Q^t = Q^{i_Q}(m_A^{t-1}, x_Q, h_Q^{t-1})$ \\
            $m_A^t, h_A^t = A^{i_A}(m_Q^{t-1}, x_A, h_A^{t-1})$ \\
        }
        $\hat{u} = U^{i_Q}(x_Q, h_Q^T)$ \\
        Policy gradient update \wrt \emph{both} \qbot{} and \abot{} parameters
    }
    \textcolor{blue}{
    \If{$e~\mathrm{mod}~E = 0$}{
        Sample replacement set $B$ under policy $\pi$ and
        re-initialize all agents in $B$
    }}
}
\Return all \qbot{}s and \abot{}s.
\end{algorithm}

\vspace{-10pt}
\xhdr{Populations of Agents.} We consider a population of \qbot{}s $\{Q^1, \ldots, Q^{N_Q}\}$ and
a population of \abot{}s $\{A^1, \ldots, A^{N_A}\}$ with each agent having a different set of parameters.
At each iteration during learning, we sample a random \qbot-\abot pair to interact and receive updates -- \ie the red line (2) in \algref{alg:train_aug}. As any \qbot may be made to communicate with any \abot, there is pressure for the population to adopt a unified language. Likewise, when an agent is reinitialized it will receive positive reward much more quickly when it happens to use language that its conversational partners understand. Furthermore, `compressible' languages that are easier to learn will result in greater reward for the population in the face of periodic re-initialization of agents. 

Introducing multiple agents may in itself add compressibility pressure and improve generalizations even without replacement \citep{raviv18}. Agents in a population have to model minor linguistic differences between conversational partners given the same memory capacity. Further, each agent provides another potential language variation that can be
mimicked and perpetuated--increasing language diversity early in training.
We examine these effects through no-replacement baselines, but find that generational pressure where
some agents know less than others can also be important for compositionality in our setting.

\xhdr{Replacement.}
In order to create a notion of `generations' we replace agents periodically.
Let $\pi$ be a replacement strategy, returning a subset of the population. Every $E$ epochs, we call $\pi$ and reinitialize the parameters and optimizers for the returned agents (blue lines 9-10 in \algref{alg:train_aug}).  We investigate three settings of $\pi$ (see appendix A.2 for more details):

\begin{compactitem}[\hspace{3pt}--]
    \setcounter{enumi}{1}
    \item \textbf{Uniform Random.} Sample an \abot and \qbot from uniform random distributions.
    \item \textbf{Epsilon Greedy.}
    With probability $1{-}\varepsilon$ replace the \abot and \qbot with the lowest validation accuracy. We use $\varepsilon = 0.2$ in our experiments.
    \item \textbf{Oldest.} Replace the oldest \abot{} and \qbot{},
    breaking ties with uniform random sampling.
\end{compactitem}

\section{Experimental Setting} \label{sec:experiments}
\label{sec:emnlp_details}

\paragraph{Experimental Setting.}
We evaluate on both our \ourdata dataset and the \emnlpdata dataset from \citet{emnlp} (see appendix A.1), as described in \secref{sec:approach_dialog}. All results are reported as means and variances computed from a total of 16 trials (four random seeds each with 4-way cross-validation). We report accuracy based on \qbot getting both elements of the task correct -- corresponding to the more restrictive ``Both'' setting from \cite{emnlp}.

\citet{emnlp} examined a series of increasingly restrictive settings in order to study conditions under which compositionality emerges. The primary variables are whether \abot has memory (ablated by setting $h_A^t {=} 0$) and the vocabulary sizes $\mathcal{V}_Q$ and $\mathcal{V}_A$ for \qbot and \abot respectively. For comparison we also evaluate in these settings: \textbf{Minimal Vocab} ( $V_Q {=} 3$, $V_A {=} 4$).
\textbf{Memoryless + Minimal Vocab} ($V_Q {=} 3$, $V_A {=} 4$, $h_A^t {=} 0$), \textbf{Overcomplete} ($V_Q {=} V_A {=} 64$). We also introduce \textbf{Memoryless + Overcomplete} ($V_Q {=} V_A {=} 64$, $h_A^t {=} 0$) to complete the cross product of settings and examine the role of memory restriction in overcomplete vocabularies.

The Memoryless + Minimal Vocabulary setting results in the best compositional generalization; however, this is an extreme setting -- requiring not only that the minimum number of groundable symbols be known but also that \abot{} not be able to remember it's previous utterance. While we do report these settings and see quite large performance gains due to cultural transmission, we are mainly interested in the more realistic Overcomplete setting where a large pool of possible tokens is provided and both dialog agents have memory.

\paragraph{Model and Training Details.}
Our \abot{}s and \qbot{}s have the same architectur
as in \citet{emnlp}.
All agents are trained with $E = 25000$, a
batch size of 1000,
\footnote{
All 384 instances (64 objects $\times$ 6 tasks) fit in 1 batch.}
and the Adam~\citep{adam}
optimizer (one per bot) with learning rate 0.01.
In the Multi Agent setting we use $N_A = N_Q = 5$.
We stop training after 8 generations (199000 epochs Multi Agent; 39000 epochs Single Agent).
This differs from \citet{emnlp}, which stopped
once train accuracy reached 100\%.
Further, we do not mine negatives.

\paragraph{Baselines.} These help isolate the effects of our approach. 

\begin{compactitem}[\hspace{3pt}--]
\item \textbf{Single Agent Populations.} We ablate the effect of multi-agent populations by training individual \abot-\qbot pairs (\ie populations with $N_A=N_Q=1$). We apply the \emph{uniform random} (either \abot or \qbot at random) and \emph{oldest} (alternating between \abot and \qbot) replacement strategies to these agents; however, the \emph{epsilon greedy} strategy is not well-defined here. In this setting we decrease $E$ from 25000 to 5000 to keep the average number of gradient updates for each agent constant with respect to the multi-agent experiments.
\item \textbf{No Replacement.} We also consider the effect of replacing no agents at all, but still allowing the agents to train for the full 199,000 (39,000) epochs. Improvement over this baseline shows the gains from our replacement strategy under identical computational budgets.
\end{compactitem}

\noindent
The code used to implement all experiments is available at \href{https://github.com/mcogswell/evolang}{https://github.com/mcogswell/evolang}.


\begin{figure*}[t]
\centering
\vspace{-20pt}
\includegraphics[width=\textwidth]{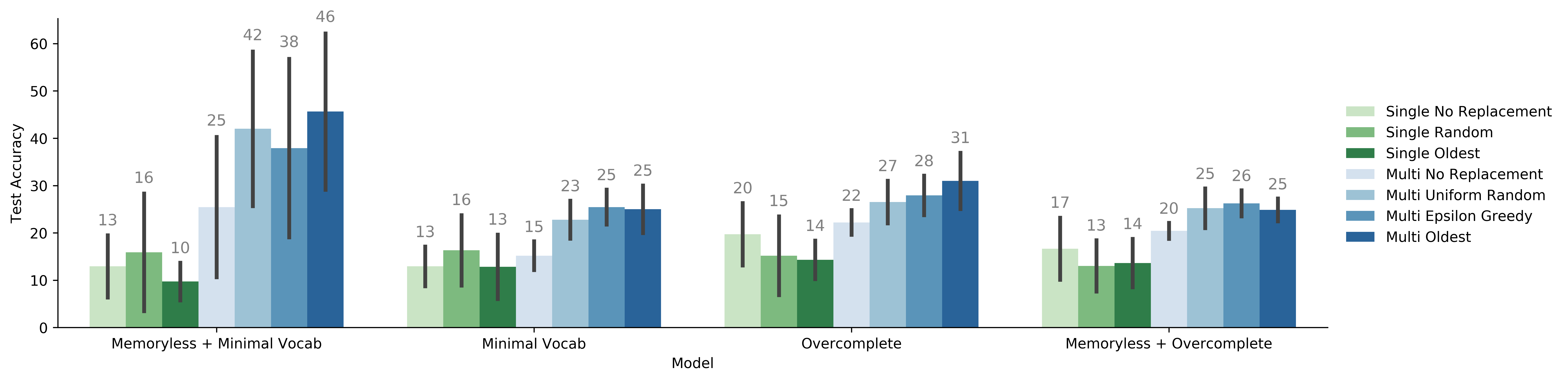}\\[-8pt]
\caption{
Test set accuracies (with standard deviations) are reported
against our new harder dataset using models similar to those in \protect\cite{emnlp}.
Our variations on cultural transmission (darker blue bars) outperform the
baselines without cultural transmission.}
\label{fig:main_results}
\vspace{-10pt}
\end{figure*}

\section{Results and Analysis}

\subsection{Impact of Cultural Transmission on Compositional Generalization}
Results with standard deviations against our harder dataset are reported in \figref{fig:main_results}.
We compared methods and models using dependent paired t-tests and reported the resulting p-values in Section A.4 of the appendix.
Result on the original Task \& Talk dataset are in Section A.1 of the appendix.

\xhdr{Cultural transmission induces compositionality.}
Our main result is that cultural transmission approaches
outperform baselines without cultural transmission.
This can be seen by noting that for each model type in \figref{fig:main_results},
the 3 darker blue bars (Multi Agent Replacement approaches) are largest.
After running a dependent paired t-test against all pairs of baselines
and cultural transmission approaches we find a meaningful difference
in all cases ($p \le 0.05$).
This is strong support for our claim that our version of cultural
transmission encourages compositional language because it causes
better generalization to novel compositions of attributes.

Next we go on to discuss some additional trends we hope
the community will find useful.

\xhdr{Population dynamics without replacement usually lead to some compositionality.}
The \emph{Multi Agent No Replacement} policies usually outperform
than the \emph{Single Agent No Replacement} policies, though the difference isn't
very significant in the except in the \emph{Overcomplete} and \emph{Minimal Vocab} settings.
This agrees with recent work from evolutionary linguistics,
where multiple agents can lead to compositionality without
generational transmission~\cite{raviv18}.

\xhdr{Variations in replacement strategy tend to not affect performance.}
The \emph{Multi Agent Uniform Random/Epsilon Greedy/Oldest} replacement strategies are not largely or consistently different from one another across model variations.
This suggests that while some agent replacement needs to occur, it is not
critical whether agents with worse language are replaced or whether there is a pool
of similarly typed agents to remember knowledge lost from older generations.
The main factor is that new agents learn in the presence of
others who already know a language.

\xhdr{Cultural transmission is complementary with other
factors that encourage compositionality.}
As in \citet{emnlp}, we find the \emph{Memoryless + Small Vocab}
model is clearly the best.
This agrees with factors noted elsewhere~\cite{emnlp,mordatch_abbeel,nowak00}
and shows how many different factors can affect the emergence of compositionality.

\xhdr{Removing memory makes only minor differences.}
Removing memory makes no difference (negative or positive) in Single Agent settings, but it can
have a relatively small effect in Multi Agent settings, helping \emph{Small Vocab} models and
hurting \emph{Overcomplete} models.
While our approach is complementary with minimizing vocab size to increase compositionality,
its makes memory removal less useful.
As the \emph{Memoryless + Overcomplete} setting has not been reported before,
these results suggest that the relationship between inter-round memory
and compositionality is not clear.

Overall, these results show that adding cultural transmission to neural dialog agents
improves the compositional generalization of the languages learned by
those agents in a way complementary to other priors. It thereby shows
how to transfer the cultural transmission principle from
evolutionary linguistics to deep learning.
\subsection{Is Generational Transmission Occurring?} \label{sec:lang_dist}



Because it is implicit, cultural transmission may not actually
be occurring; improvements may be from other sources.
How can we measure cultural transmission?
We focus on \abot{}s and take a simple approach. We assume that if two \abot{}s `speak
the same language' then that language was culturally
transmitted. There is a combinatorial explosion of possible
languages that could refer to all the objects of interest,
so if the words that refer to the same object for two agents
are the same then they were very likely transmitted from the
other agents, rather than similar languages
emerging from scratch just by chance.
This leads to a simple approach: consider pairs of bots
and see if they say similar things in the same context.
If they do, then their language was likely transmitted.

\begin{figure*}[t]
\centering
\vspace{-20pt}
\includegraphics[width=\textwidth,trim={0pt 7pt 0pt 7pt},clip]{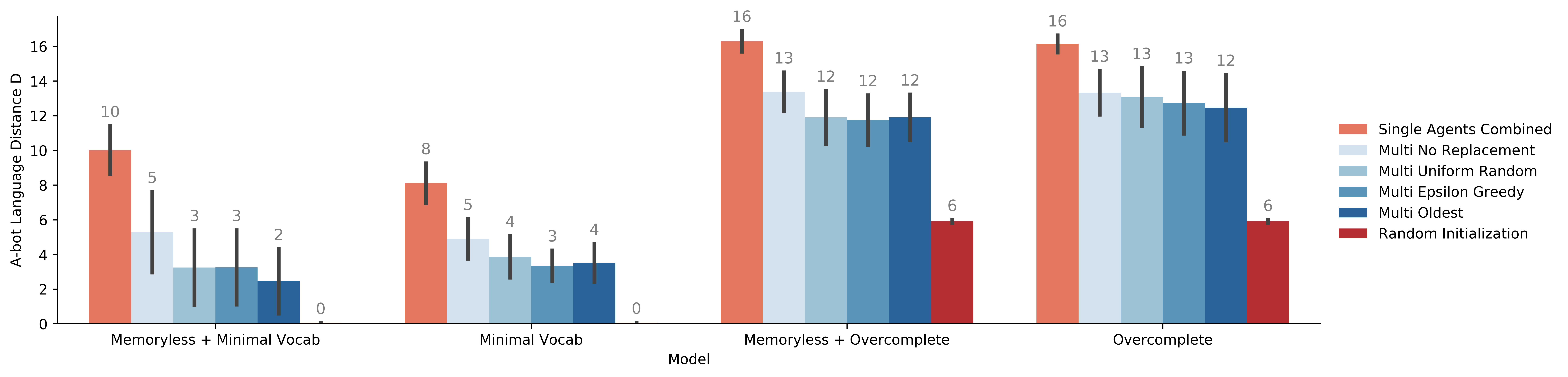}\\[-6pt]
\caption{
Do bots in a population learn similar languages?
On the y-axis (eq. \eqref{eq:D}) lower values indicate more similar language.
Bots from our method speak similar languages, but independently evolved agents do not.
Thus our implicit procedure induces cultural transmission.}
\label{fig:emnlp_lang_dists}
\vspace{-10pt}
\end{figure*}

More formally, consider the distribution of tokens \abot{} $A^i$
might use to describe its object $x_A$ when talking to \qbot{} $Q^k$:
$p_{k,i}(m_A^t | x_A)$ or $p_{k,i}$ for short.
We want to know how similar $A^i$'s language is to that of
another \abot{} $A^j$. We'll start by comparing those two distributions
by computing the KL divergence between them and then taking an
average over context (objects, \qbot{}s, and dialog rounds)
to get our pairwise agent language similarity metric $D_{ij}$:
\begin{align}
D_{ij} &= 
\hat{E}_{x_A,k,t}
\left[
        D_{KL}\left(p_{k,i}(m_A^t | x_A), p_{k,j}(m_A^t | x_A)\right)
\right]
\label{eq:Dij}
\end{align}
Taking another average, this time over all pairs of bots (and also random seeds
and cross-val folds), gives our
final measure of language similarity reported in \figref{fig:emnlp_lang_dists}.
\begin{align}
D &= \hat{E}_{i,j~\text{s.t.}~i \ne j} \left[ D_{ij} \right]
\label{eq:D}
\end{align}
$D$ is smaller the more similar language is between bots.
Note that even though $D_{ij}$ is not symmetric (because KL divergence is not),
$D$ is symmetric because it averages over both directions of pairs.

We compute $D$ by sampling an empirical distribution over all
messages and observations, taking 10 sample dialogues in each possible
test state $(x_A, x_Q)$ of the world using the final populations of agents as in \figref{fig:main_results}.
Note that this metric applies to a group of agents, so we measure it for
only the \emph{Multi Agent} settings, including two new baselines colored red in \figref{fig:emnlp_lang_dists}.
The \emph{Single Agents Combined} baseline trains 4 \emph{Single Agent No Replacement}
models independently then puts them together and computes $D$ for that group.
These agents only speak similar languages by chance, so $D$ is high.
The \emph{Random Initialization} baseline evaluates language similarity using
newly initialized models. These agents have about a uniform distribution over
words at every utterance, so their languages are both very similar and useless.
For each model these baselines act like practical (not strict)
upper and lower bounds on $D$, respectively.

\figref{fig:emnlp_lang_dists} shows this language dissimilarity metric for all our settings. As we expect, the paired \emph{Single Agents} are highly dissimilar compared to agents
from \emph{Multi Agent} populations. Further, all the replacement strategies result in increased language similarity---although
the degree of this effect seems dependent on vocabulary setting. This provides some evidence that cultural transmission is occurring in \emph{Multi Agent} settings and is encouraged by the replacement strategy in our approach.
While all \emph{Multi Agent} settings resulted in language transmission, our replacement strategies results in more compositional languages due to repeated teaching of new generations of agents.

\begin{figure}[!h]
    \centering
    \begin{subfigure}[c]{0.30\textwidth}
        \includegraphics[width=1\textwidth]{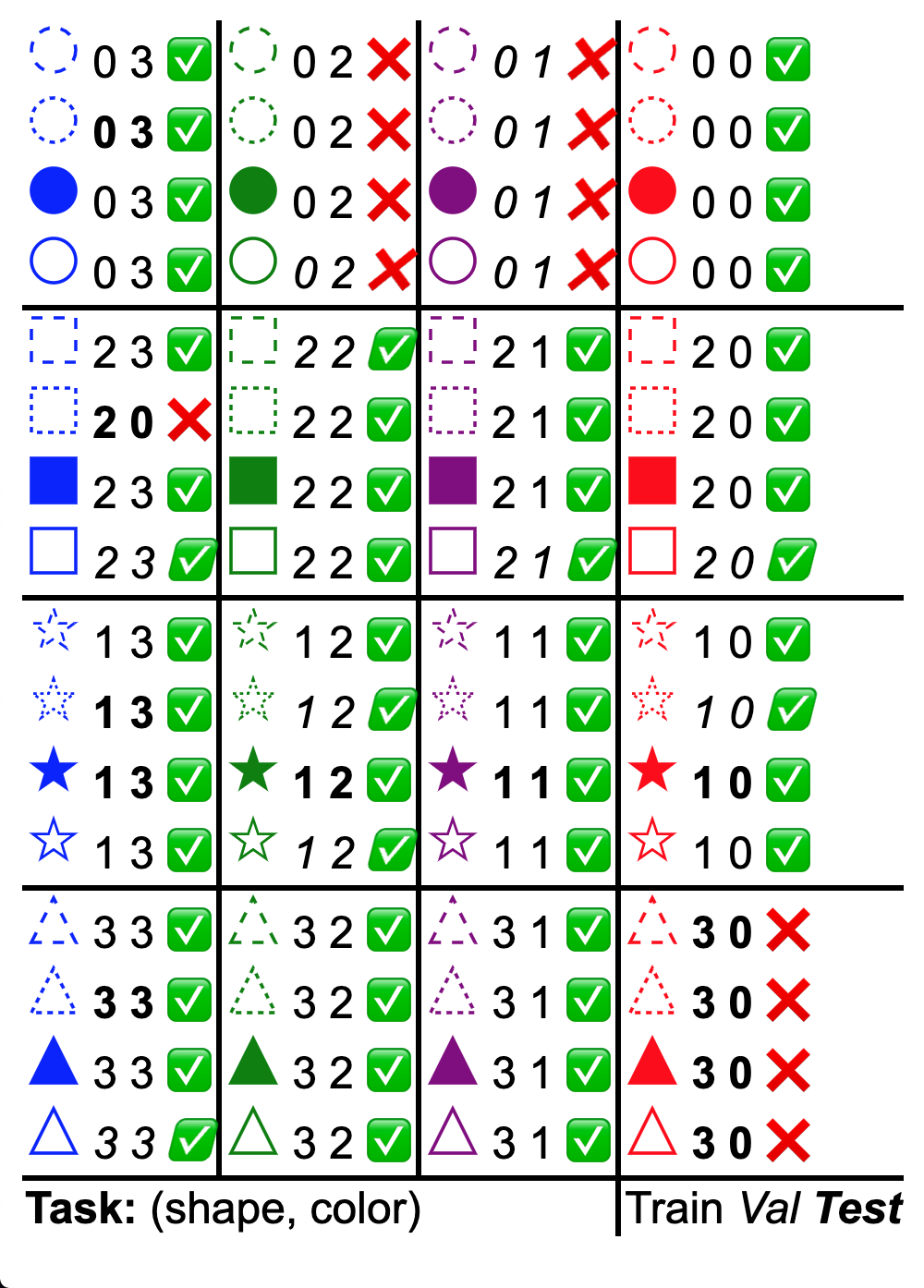}
    \end{subfigure}
    
    \caption{All conversations between \qbot{} 4 and \qbot{} 3 for the \texttt{(shape, color)} task.
    These bots were trained in the \emph{Multi Agent Oldest} setting.
    The figure shows \abot{}'s utterances for each object and whether or not
    \qbot{} guessed the object correctly, as described in \secref{sec:qual}.
    This language is compositional because each token refers to a color or shape.
    }
    \label{fig:lang_vis}
    
\end{figure}

\subsection{Visualizing Emergent Languages} \label{sec:qual}

In this section we visualize the language learned by a pair of
bots to show its compositionality.
In the appendix we compare these bots to others at different
stages of learning and from earlier generations to help understand
how the language developed over generations.

Figure \ref{fig:lang_vis} shows all 64
conversations between \qbot{} 4 and \abot{} 3 for the \texttt{(shape, color)} task.
These bots are from the 8th generation of the \emph{Multi Agent Oldest} setting.

To interpret the visualization, start by looking at only the dashed blue circle in the top left.
To the right of it are the two tokens ``0'' and ``3'', which are the words
\abot{} used to describe the object in the two dialog rounds.
The green check one more step to the right indicates that \qbot{} was able to guess the
\texttt{circle blue} from these tokens.
Now look at all 4 blue circles in the top left grid cells. Only \texttt{shape} and 
\texttt{color} matter for this task, so \abot{} uttered ``0 3'' for every blue
circle, appropriately ignoring style (\ie, dashed, dotted, filled, or solid).

By looking at the entire visualization with its 4x4 grid
delineated by black separators we can see that the language is indeed compositional.
Rows of the 4x4 grid group objects by shape
and columns group objects by color.
This makes it convenient to qualitatively evaluate language compositionality with respect
to the \texttt{(shape, color)} task.
If \abot{}'s language is compositional then it should use one token to indicate row / shape
and one token for column / color.

Looking at the first row, \abot{}'s first utterance is always ``0'',
but is not ``0'' anywhere else, so when ``0'' is uttered first it means \texttt{circle}.
Similarly, \abot{}'s second utterance is always ``3'' in the first column, so
``3'' means \texttt{blue}. Continuing with this analysis we find each character has meaning:
(0=\texttt{circle}, 2=\texttt{square}, 1=\texttt{star}, 3=\texttt{triangle}), and (3=\texttt{blue}, 2=\texttt{green}, 1=\texttt{purple}, 0=\texttt{red}).
Individual symbols have meaning, so the language is compositional.
\section{Related work} \label{sec:relwork}


\paragraph{Language Evolution Causes Structure.}
Researchers have spent decades studying how unique properties of human
language like compositionality could have emerged.
There is general agreement that 
people acquire language using a combination of innate cognitive
capacity and learning from other language speakers (cultural 
transmission), with the degree of each being widely 
disputed~\cite{perfors02,pinker1990}.
Both innate cognitive capacity and specific modern human languages like English
co-evolved~\cite{briscoe00} via biological~\cite{pinker1990} and 
cultural~\cite{tomasello99,smith06} evolution, respectively.

In particular, explanations of how the cultural evolution of languages 
could cause structure like compositionality are in 
abundance~\cite{nowak_krakauer99,nowak00,smith_kirby03,brighton02,vogt05,kirby_survey14,spike_reqs}.
An important piece of the explanation of linguistic structure is the iterated learning
model~\cite{kirby_survey14,kirby01,kirby08} used to motivate our approach. Indeed
it shows that cultural transmission causes structure in
computational~\cite{kirby01,kirby02,kirby_survey03,smith_kirby03} and 
human~\cite{kirby08,cornish_kirby09,kirby_survey10} experiments.
Even though cultural transmission may aid the emergence of compositionality,
recent results in evolutionary linguistics~\cite{raviv18} and deep learning~\cite{emnlp,mordatch_abbeel}
also emphasize other factors.

While existing work in deep learning has focused on biases that
encourage compositionality, it has not considered settings where
language is permitted to evolve over generations of agents.
We have shown such an approach is viable and even complementary with other approaches.

\paragraph{Language Emergence in Deep Learning.}
Recent work in deep learning has increasingly focused on multi-agent
environments where deep agents learn to accomplish goals
(possibly cooperative or competitive) by interacting appropriately
with the environment and each other.
Some of this work has shown that deep agents will develop their
own language where none exists initially if driven by a
task which requires communication~\cite{foerster16,fergus16,baroni16}.
Most relevant is work which focuses on conditions under which
\emph{compositional} language emerges as deep agents learn to cooperate~\cite{mordatch_abbeel,emnlp}.
Both \citet{mordatch_abbeel} and \citet{emnlp} find that limiting
the vocabulary size so that there aren't too many more words than
there are objects to refer to encourages compositionality,
which follows earlier results in evolutionary linguistics~\cite{nowak00}.
Follow up work has continued to investigate the emergence of
compositional language among neural agents, mainly focusing
on perceptual as opposed to symbolic input and how the structure of
the input relates to the tendency for compositional language
to emerge~\cite{choi18,havrylov17,lazaridou18}.
Other work has shown
that Multi Agent interaction leads to better emergent translation~\cite{lee18emerge},
but it does not measure compositionality.

\paragraph{Cultural Evolution and Neural Nets.}
Somewhat recently, \citet{bengio12_culture} suggested
that culturally transmitted ideas may help in escaping from local minima.
Experiments in \citet{knowledge_matters} support this idea by
showing that supervision of
intermediate representations allows a more complex toy task to be learned.
Unlike our work, these experiments use direct supervision provided
by the designed environment rather than indirect and implicit supervision
provided by other agents.

Two concurrent works examine the role of periodic agent replacement on language
emergence -- albeit in different environments.
In \citet{ease_of_teaching} replacement is used to encourage languages to be easy
to teach, and this in turn causes compositionality.
In \citet{deep_coev} neural language is transmitted through a bottleneck
caused by replacement. The resulting language has increased efficiency and effectiveness,
with further results showing that co-evolving the agents themselves with the language
amplifies the effect. Both of these works support our central observations.

\section{Conclusion} \label{sec:conclusion}

In this work we investigated cultural transmission in deep neural dialog agents,
applying it to language emergence. The evolutionary linguistics community
has long used cultural transmission to explain how compositional languages
could have emerged. The deep learning community, having recently become
interested in language emergence, has not investigated that link
until now. Instead of explicit
models of cultural transmission familiar in evolutionary linguistics,
we favor an implicit model where language is transmitted from generation
to generation only because it helps agents achieve their goals.
We show that this does indeed cause cultural transmission and compositionality.

\paragraph{Future work.}
While our work used an implicit version of cultural transmission,
we are interested in the effect of explicit versions of cultural transmission
on language structure.
Cultural transmission may also provide an appropriate prior
for neural representations of non-language information.

\subsubsection{Acknowledgements}

We would like to thank Satwik Kottur for code and comments as well as Karan Desai for additional code. We would also like to thank Douwe Kiela, Diane Bouchacourt, Sainbayar Sukhbaatar, Marco Baroni, and Erik Wijmans for comments on earlier versions of this paper.

The Georgia Tech effort was supported in part by NSF, AFRL, DARPA, ONR YIPs, ARO PECASE, and Amazon. The views and conclusions contained herein are those of the authors and should not be interpreted as necessarily representing the official policies or endorsements, either expressed or implied, of the U.S. Government, or any sponsor.

\bibliographystyle{named}
\bibliography{main}

\clearpage

\appendix
\section{Appendix}
\subsection{Results on Novel Instance Dataset of \protect\cite{emnlp}}
\label{sec:emnlp_dataset}

In section 2 of the main paper we discuss the difference between
the \emnlpdata and \ourdata datasets.
Our \ourdata dataset is a more difficult compositional split,
than the one in \citet{emnlp}. For comparison, in this section we
train and evaluate our models on the \emnlpdata from \citet{emnlp} to
show that our approach still improves compositionality in this
setting and to show that our new dataset is indeed more difficult.

\begin{figure*}[h]
\centering
\includegraphics[width=\textwidth]{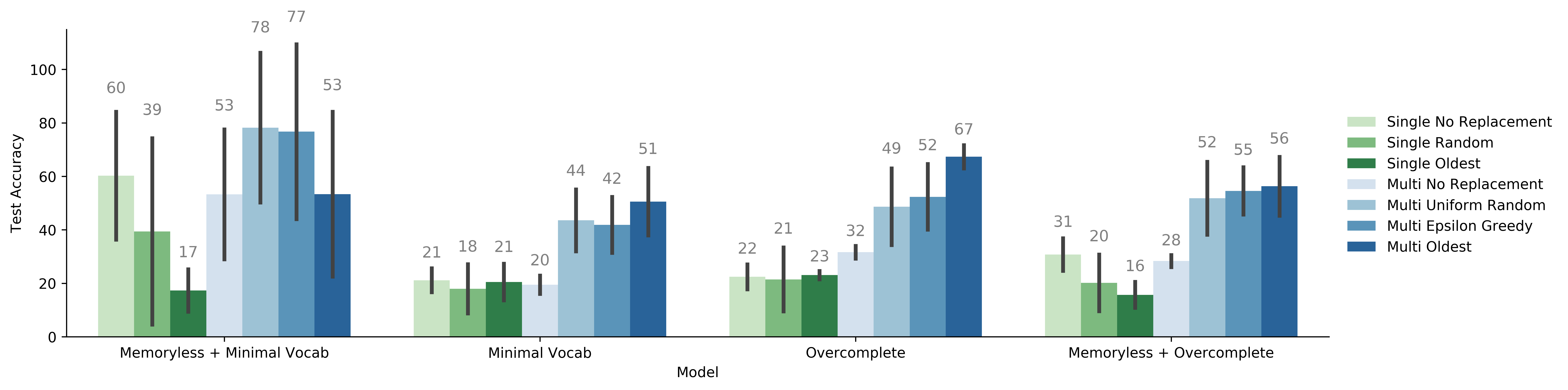}\\[-8pt]
\caption{
Test set accuracies (with standard deviations) are reported
by training and evaluating the same models as in our main results (figure 2 main paper)
against the dataset from \protect\cite{emnlp}.
These results do not perform cross-validation, following \protect\cite{emnlp}.
They only vary across 4 different random seeds.
See \secref{sec:emnlp_dataset}.
}
\label{fig:emnlp_results}
\vspace{-10pt}
\end{figure*}

In \figref{fig:emnlp_results} test set accuracies (with standard deviations) are reported
by training and evaluating the same models as in our main results (figure 2 main paper)
against the dataset from \protect\cite{emnlp}.
These results do not perform cross-validation, following \protect\cite{emnlp}.
They only vary across 4 different random seeds.
Our proposed approach still outperforms models without replacement and without multiple agents.
Furthermore, by comparing the approaches from \figref{fig:emnlp_results} to figure 2 from the main paper we can
see much lower performance across the board on the \ourdata than on the \emnlpdata dataset used here.
This indicates the \emnlpdata dataset is significantly easier than our new dataset, and that
our models encourage compositionality in both settings.

\subsection{Replacement Strategies}
\label{sec:replacement}

Our approach to cultural transmission periodically replaces
agents by re-initializing them.
The approach section outlines various replacement strategies (policy $\pi$),
but does not detail their implementation.
We do so here.

These strategies depend on a number of possible inputs:
\begin{itemize}
    \item $e$ the current epoch
    \item $E$ the period of agent replacement
    \item $v^Q_i/v^A_i$ the validation accuracy of agent $i$ for \qbot{}s/\abot{}s.
    For \qbot{}s this is averaged over all potential \abot{} partners, and vice-versa for \abot{}s.
    \item $a^Q_i/a^A_i$ the age in epochs of agent $i$ for \qbot{}s/\abot{}s
\end{itemize}

Single Agent strategies are given in
\algref{alg:replace_single_random} and
\algref{alg:replace_single_alternate}.
Multi Agent strategies are given in
\algref{alg:replace_multi_random},
\algref{alg:replace_multi_eps}, and
\algref{alg:replace_multi_oldest}.
Note that Single Agent strategies always replace one agent
while Multi Agent strategies always replace one \qbot{} and one \abot{}.

\begin{algorithm}[!h]
\caption{Single Agent - Random Replacement}
\label{alg:replace_single_random}
$d \sim \calU\{0, 1\}$ \\
\If{$d = 0$}{
    \Return $\{$ \abot{} $\}$
}
\Else{
    \Return $\{$ \qbot{} $\}$
}
\end{algorithm}

\begin{algorithm}[!h]
\caption{Single Agent - Alternate Replacement}
\label{alg:replace_single_alternate}
\textbf{Input:} $e$ \\
\If{$\floor{e / E} = 0$}{
    \Return $\{$ \abot{} $\}$
}
\Else{
    \Return $\{$ \qbot{} $\}$
}
\end{algorithm}
    
\begin{algorithm}[!h]
\caption{Multi Agent - Uniform Random Replacement}
\label{alg:replace_multi_random}
$i_A \sim \calU\{1, N_A\}$ \\
$i_Q \sim \calU\{1, N_Q\}$ \\
\Return $\{$ \abot{} $i_A$, \qbot{} $i_Q$ $\}$
\end{algorithm}

\begin{algorithm}[!h]
\caption{Multi Agent - Epsilon Greedy Replacement}
\label{alg:replace_multi_eps}
\textbf{Input:} $v^Q_i \forall i$, $v^A_i \forall i$, $\varepsilon \in [0, 1)$ (usually 0.2) \\
$d \sim \calU[0, 1)$ \\
\If{$d < \varepsilon$}{
    $i_A \sim \calU\{1, N_A\}$ \\
    $i_Q \sim \calU\{1, N_Q\}$
}
\Else{
    $i_A = \argmin_i v^A_i$ \text{(unique in our experiments)} \\
    $i_Q = \argmin_i v^Q_i$ \text{(unique in our experiments)}
}
\Return $\{$ \abot{} $i_A$, \qbot{} $i_Q$ $\}$
\end{algorithm}

\begin{algorithm}[!h]
\caption{Multi Agent - Oldest Replacement}
\label{alg:replace_multi_oldest}
\textbf{Input:} $a^Q_i \forall i$, $a^V_i \forall i$ \\
$i_A = \calU\{ \argmax_i a^A_i$ \} \\
$i_Q = \calU\{ \argmax_i a^Q_i$ \} \\
\Return $\{$ \abot{} $i_A$, \qbot{} $i_Q$ $\}$
\end{algorithm}

\subsection{Visualization for Language Comparison at Dififerent Training Stages}

\begin{figure*}[h]
    \centering
    \vspace{-10pt}
    \begin{subfigure}[c]{0.3\textwidth}
        \includegraphics[width=1\textwidth]{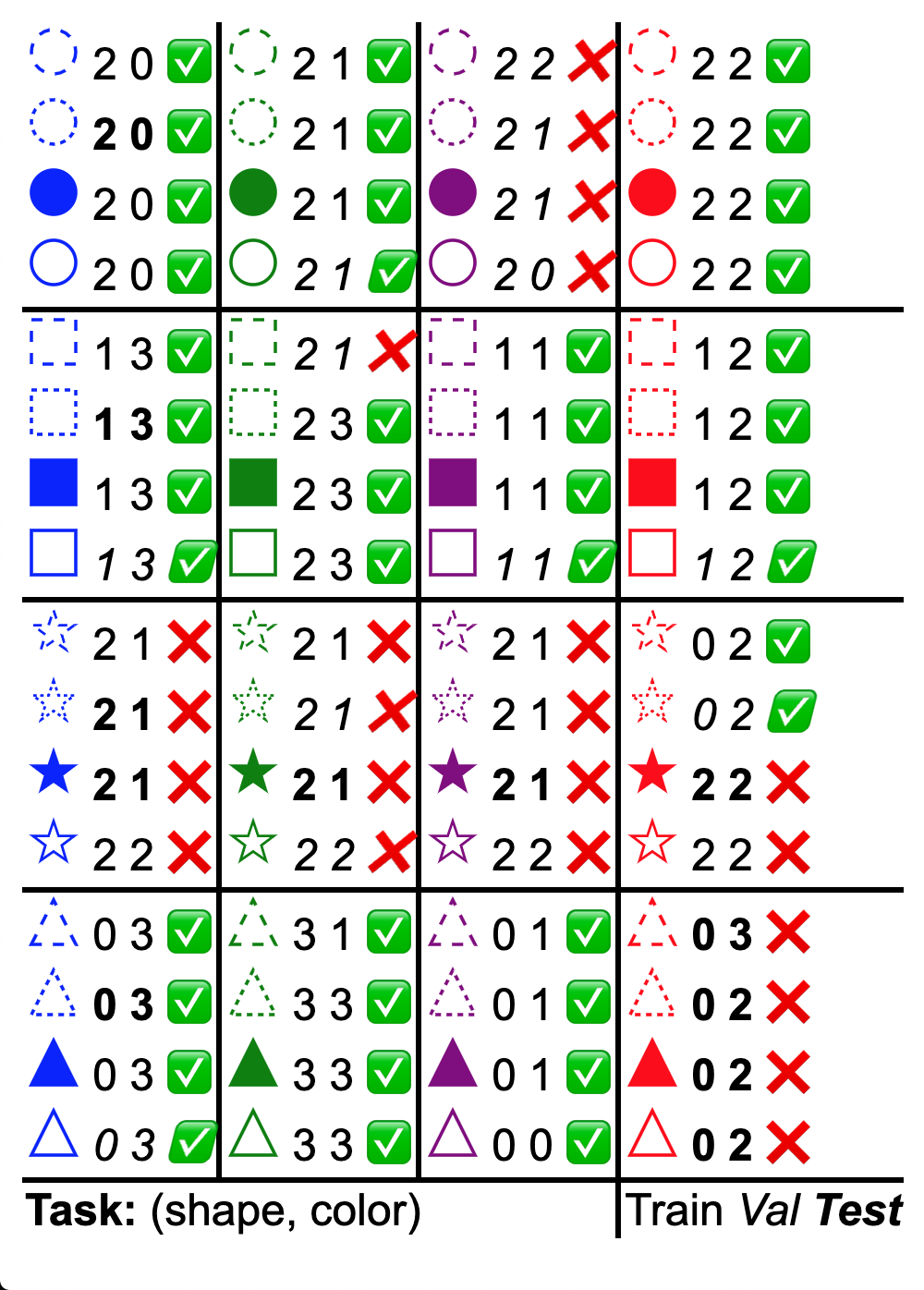}
        \caption{Gen 1 (Single) - New \abot{}}
        \label{fig:lv_single}
    \end{subfigure}
    \begin{subfigure}[c]{0.3\textwidth}
        \includegraphics[width=1\textwidth]{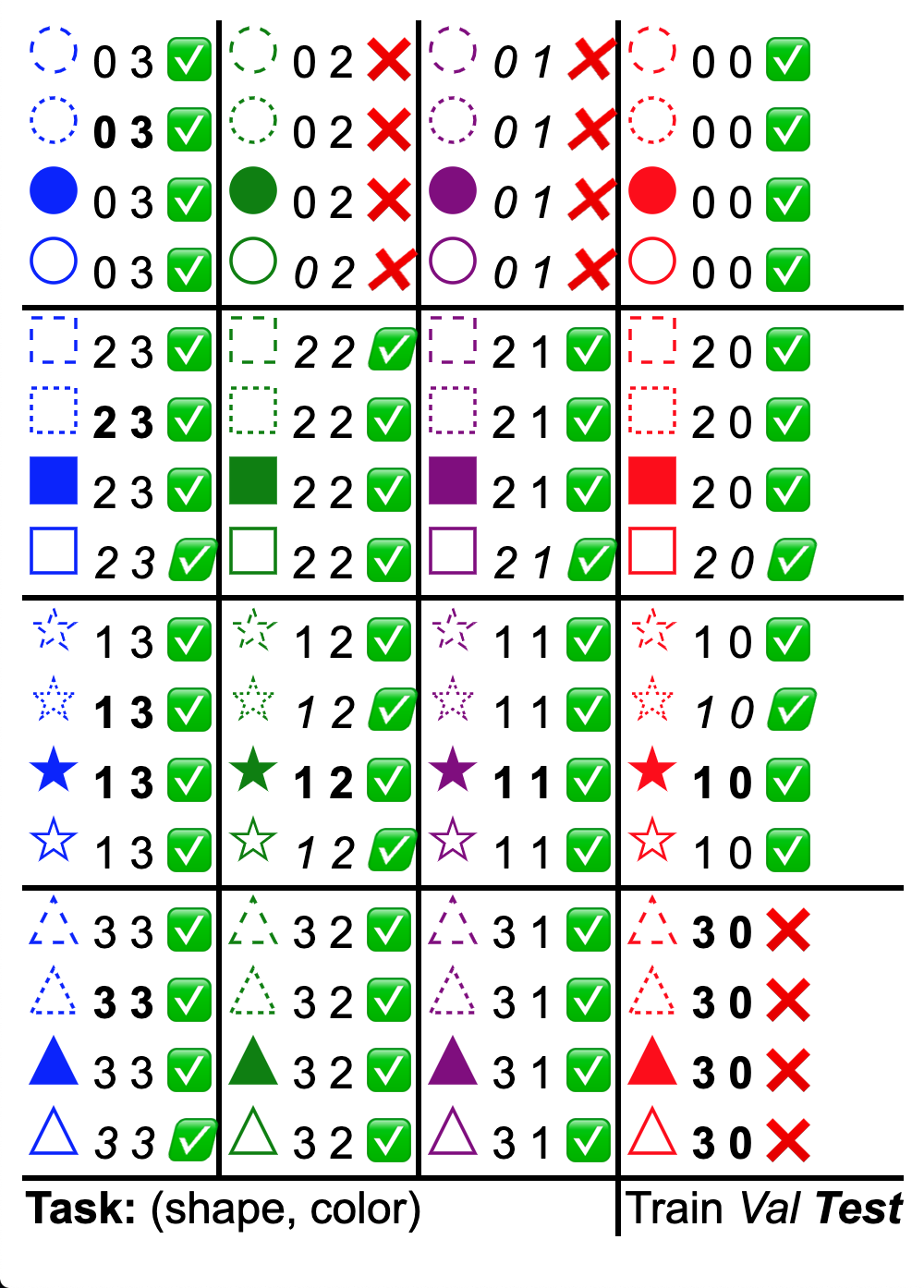}
        \caption{Gen 8 (Multi) - New \abot{}}
        \label{fig:lv_gen8_new}
    \end{subfigure}
    \begin{subfigure}[c]{0.3\textwidth}
        \includegraphics[width=1\textwidth]{images/lang_vis/multi_oldest_190000_q4_a2.png}
        \caption{Gen 8 (Multi) - Old \abot{}}
        \label{fig:lv_gen8_old}
    \end{subfigure}
    
    \caption{Each sub-figure summarizes an \abot{}'s language,
    as described in section 5.3 of the main paper. By comparing the baseline of \figref{fig:lv_single} to a similar pair of bots from our approach \figref{fig:lv_gen8_new} we can see that our approach
    encourages compositional language to emerge. Furthermore, the similarity between \figref{fig:lv_gen8_new} and \figref{fig:lv_gen8_old} suggests language is indeed transmitted in our approach.}
    \label{fig:appendix_lang_vis}
    
\end{figure*}

In this section we visualize the language learned by agents at various stages
of training to reinforce our previous conclusions and build intuition.
This builds on the visualization described in section 5.3 of the main paper,
so reference that section to individually understand the three sub-figures in \figref{fig:appendix_lang_vis}.

Each of the three sub-figures in \figref{fig:appendix_lang_vis} summarizes all of the
conversations between a particular pair of bots for the \texttt{(shape, color)} task.
From left to right: \figref{fig:lv_single} summarizes the single pair from a Single Agent No
Replacement run (3000 iterations old); \figref{fig:lv_gen8_new} summarizes dialogs between an old \qbot{}
(about 23000 iterations) and a recently re-initialized \abot{} (about 3000 iterations)
at the 8th and final generation of a Multi Oldest run;
\figref{fig:lv_gen8_old} summarizes dialogs between the same old \qbot{} as
in \figref{fig:lv_gen8_new} and an old \abot{} (13000 iterations) from the same Multi Oldest experiment.

Even though the \abot{}s in \figref{fig:lv_single} and \figref{fig:lv_gen8_new} have
trained for about\footnote{Due to the stochastic nature of our Multi Agent approach.}
the same
number of iterations, the \abot{} trained in the presence of other bots which already
know a functional language has already learned a somewhat compositional language whereas
the Single Agent \abot{} has not (\qbot{}'s gets almost all star instances wrong in \figref{fig:lv_single}, but not in \figref{fig:lv_gen8_new}).
Furthermore, by comparing the old \abot{}'s language \figref{fig:lv_gen8_old} with the new one 
\figref{fig:lv_gen8_new} we can see that they are extremely similar. They even lead to the same 
mistakes (green circles, purple circles, red triangles).
This correlation in mistakes again suggests that language is transmitted between bots, in agreement with our
previous experiments.

\subsection{Detailed Results}
\label{sec:pvalues}

In our experiments we compare models and we compare replacement strategies.
We ran dependent paired t-tests across random seeds, cross-val folds, and replacement strategies to compare models.
We ran dependent paired t-tests across random seeds, cross-val folds, and models to compare replacement strategies.
The p-values for all of these t-tests are reported here.

\begin{figure*}[!h]
    \centering
    \begin{minipage}{.4\textwidth}
        \centering
        \includegraphics[width=\linewidth]{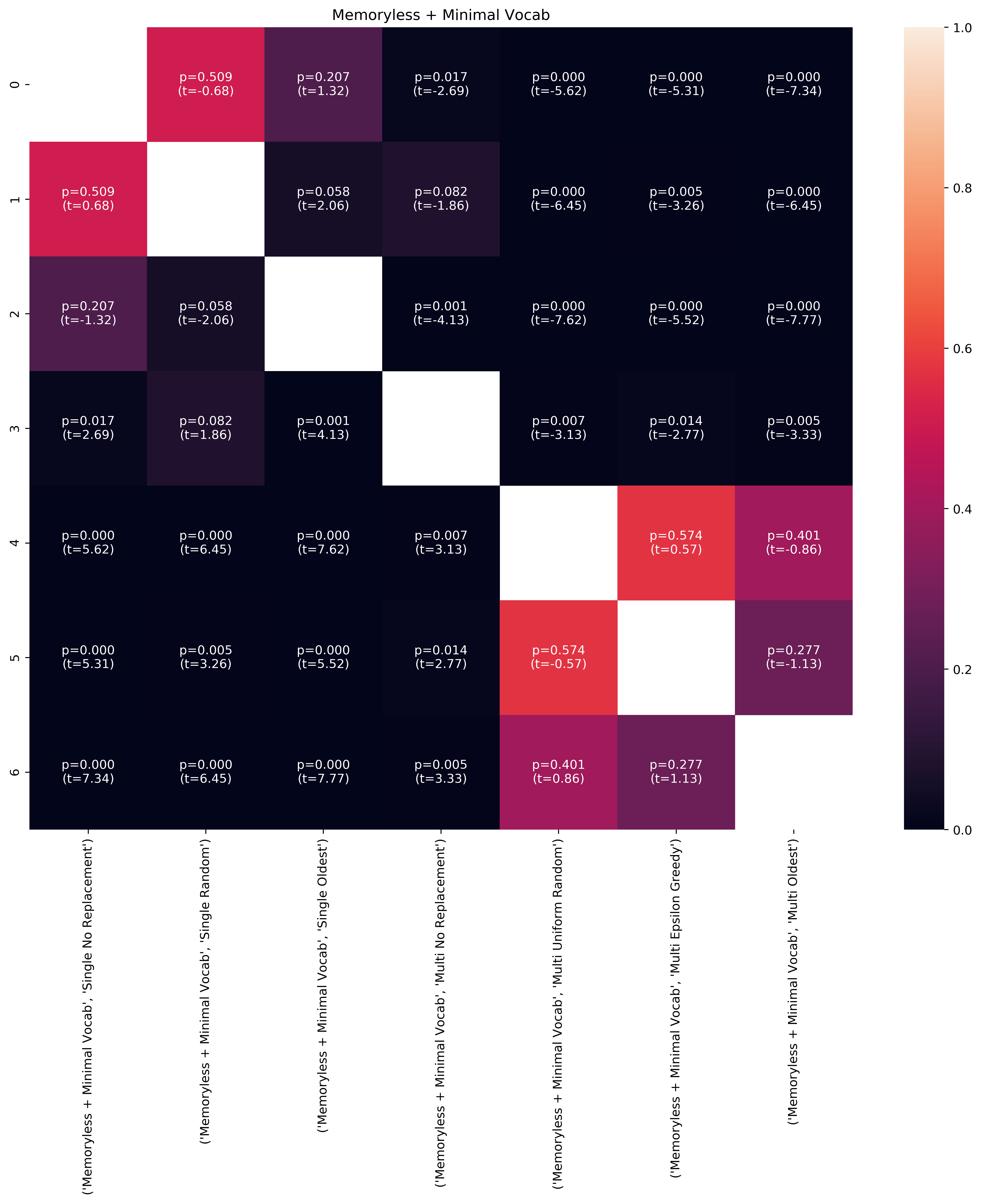}
    \end{minipage}
    \begin{minipage}{0.4\textwidth}
        \centering
        \includegraphics[width=\linewidth]{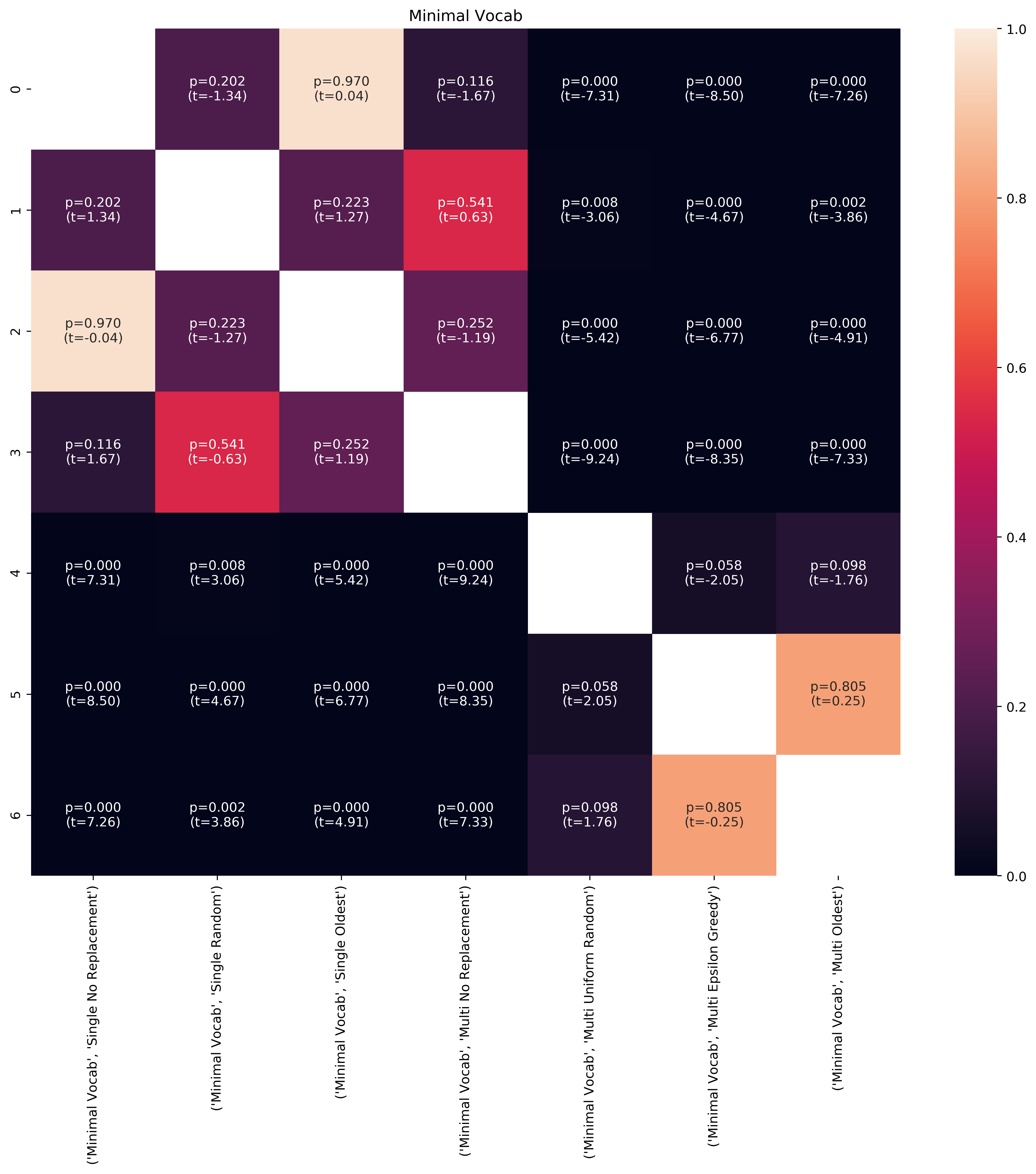}
    \end{minipage} \\
    \begin{minipage}{.4\textwidth}
        \centering
        \includegraphics[width=\linewidth]{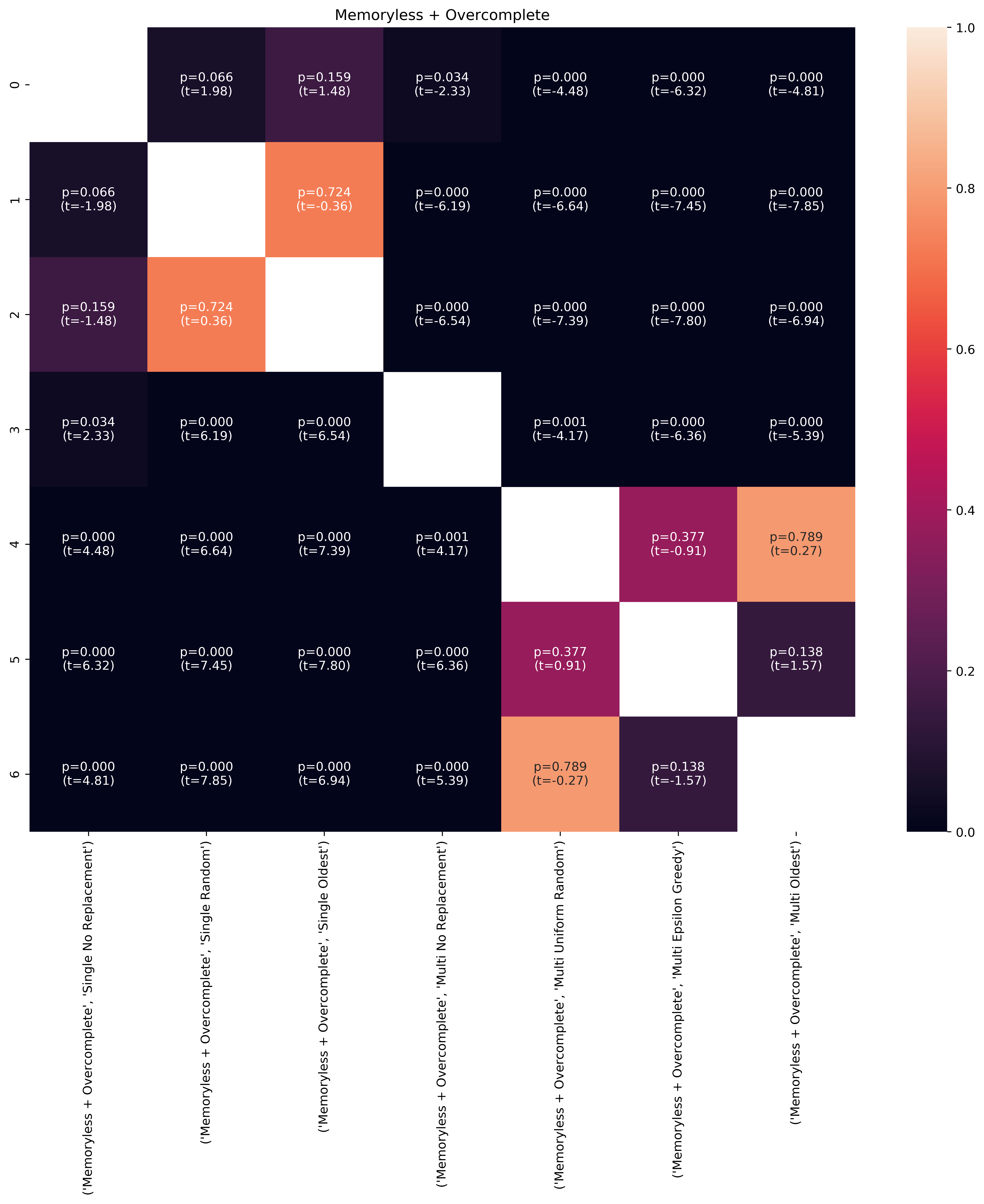}
    \end{minipage}
    \begin{minipage}{0.4\textwidth}
        \centering
        \includegraphics[width=\linewidth]{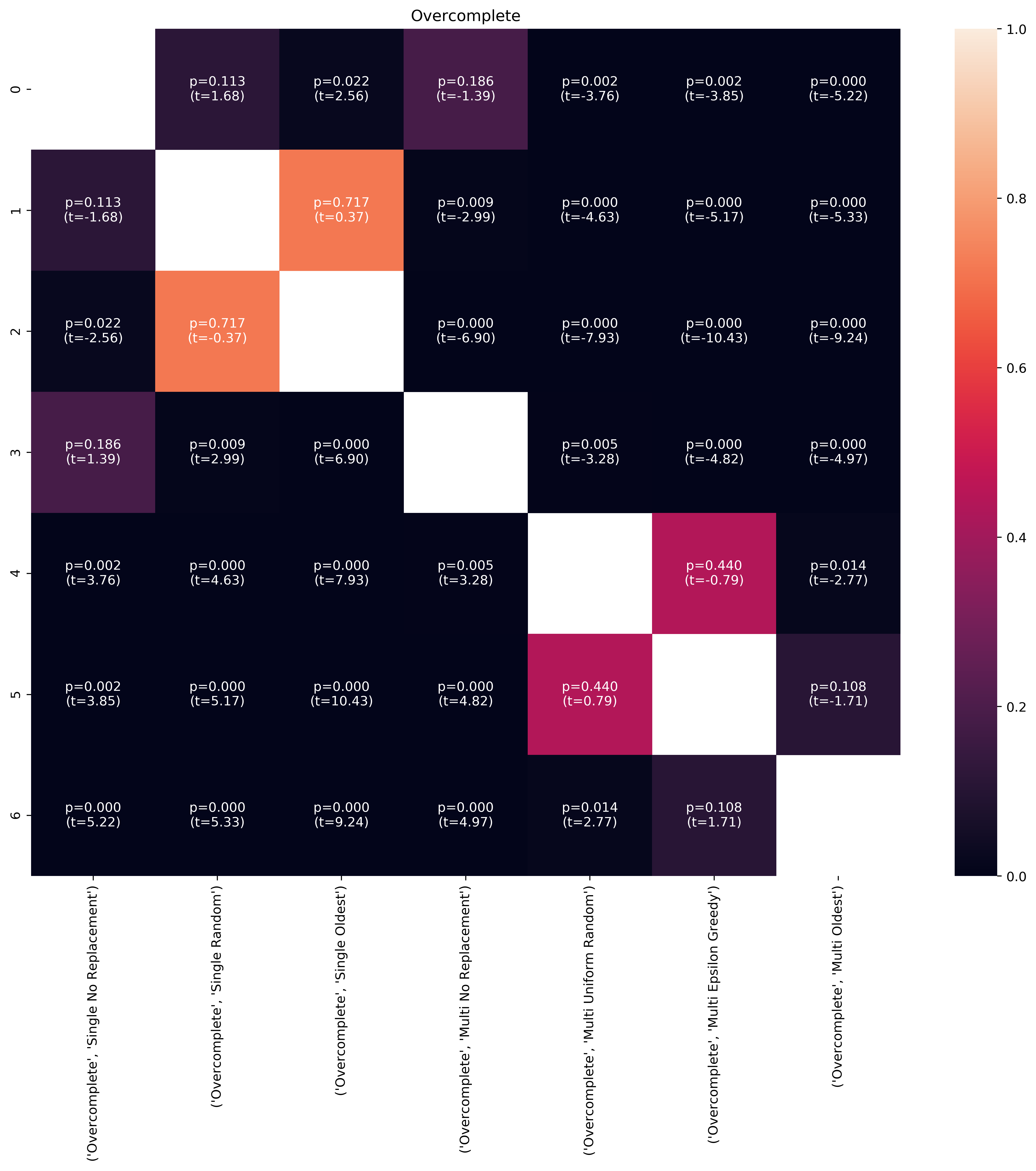}
    \end{minipage}
    \caption{Replacement strategy comparison p-values.}
    \label{fig:pvalue_model}
\end{figure*}

\begin{figure*}[!h]
    \centering
    \begin{minipage}{.35\textwidth}
        \centering
        \includegraphics[width=\linewidth]{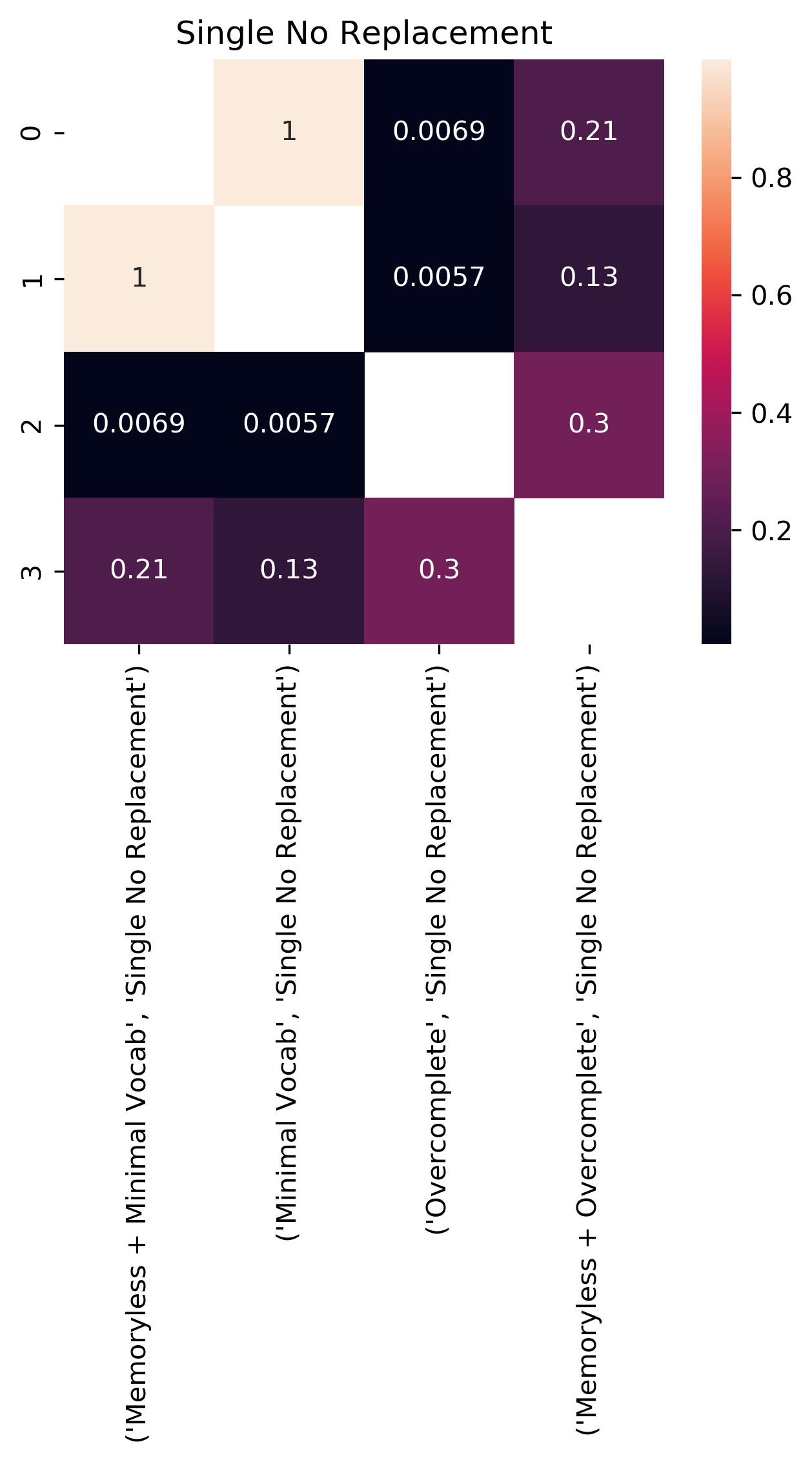}
    \end{minipage}
    \begin{minipage}{.35\textwidth}
        \centering
        \includegraphics[width=\linewidth]{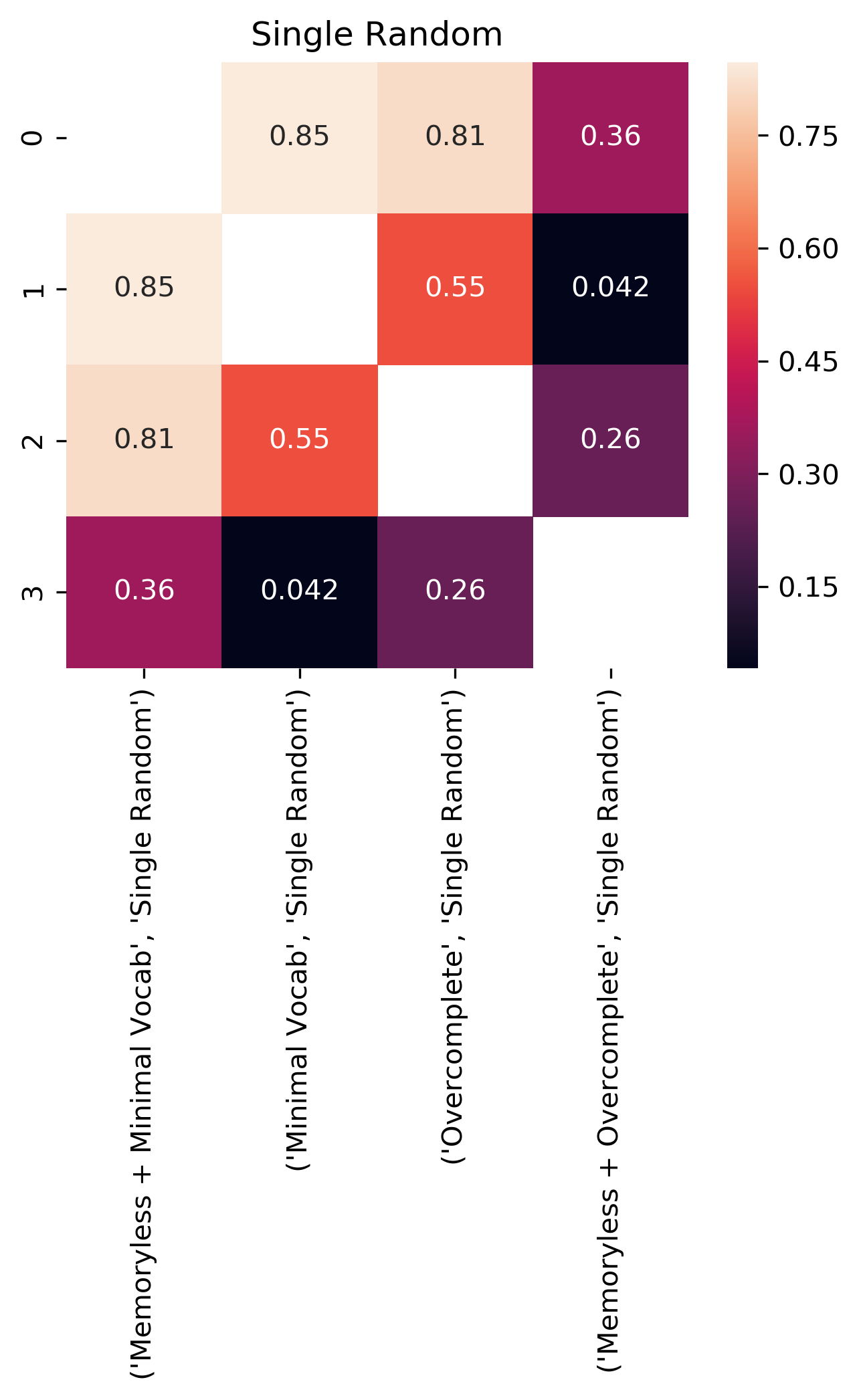}
    \end{minipage}
    \begin{minipage}{0.35\textwidth}
        \centering
        \includegraphics[width=\linewidth]{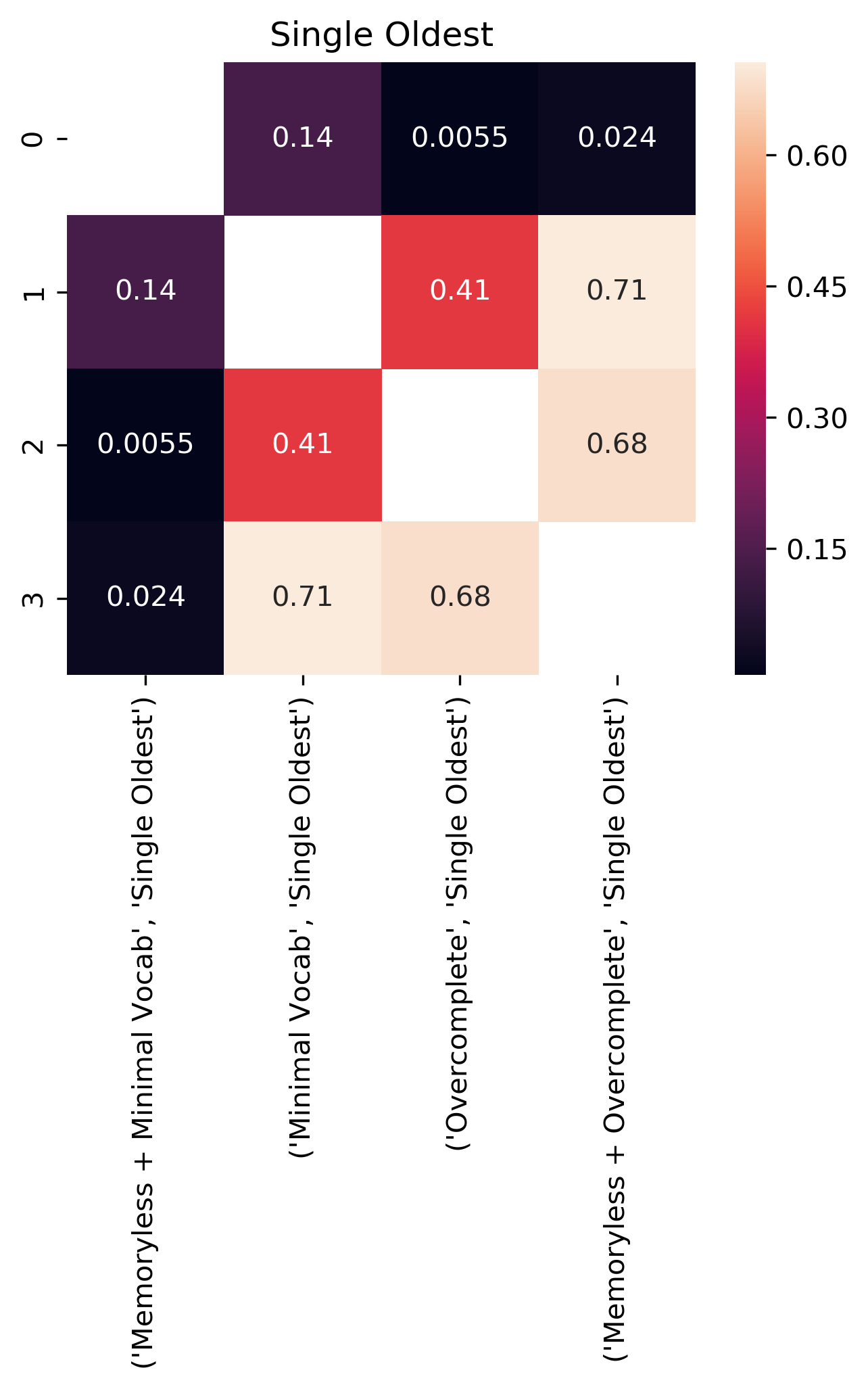}
    \end{minipage}
    \caption{Single Agent model comparison p-values.}
    \label{fig:pvalue_method_single}
\end{figure*}

\begin{figure*}[!h]
    \centering
    \begin{minipage}{.33\textwidth}
        \centering
        \includegraphics[width=\linewidth]{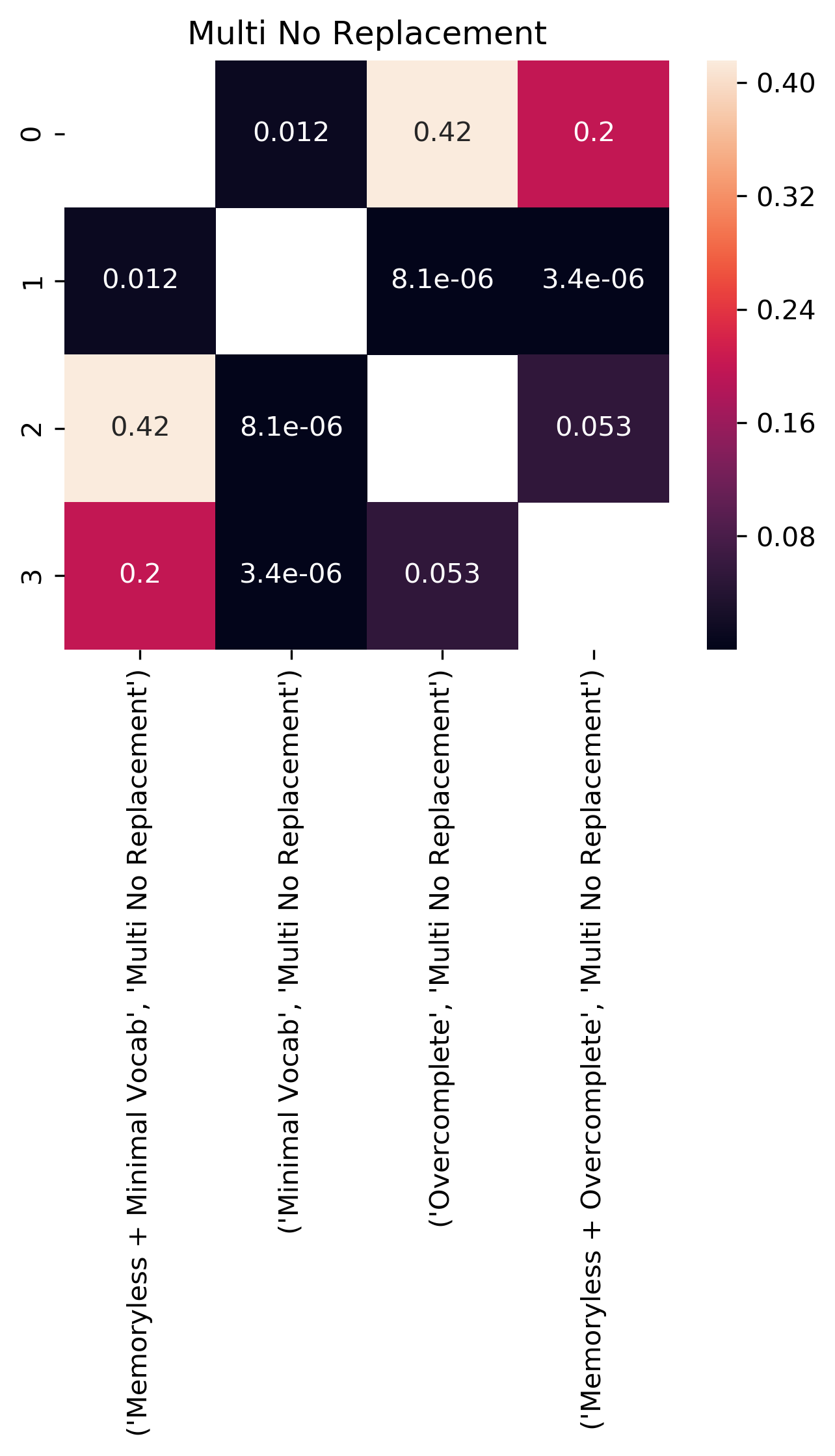}
    \end{minipage}%
    \begin{minipage}{0.33\textwidth}
        \centering
        \includegraphics[width=\linewidth]{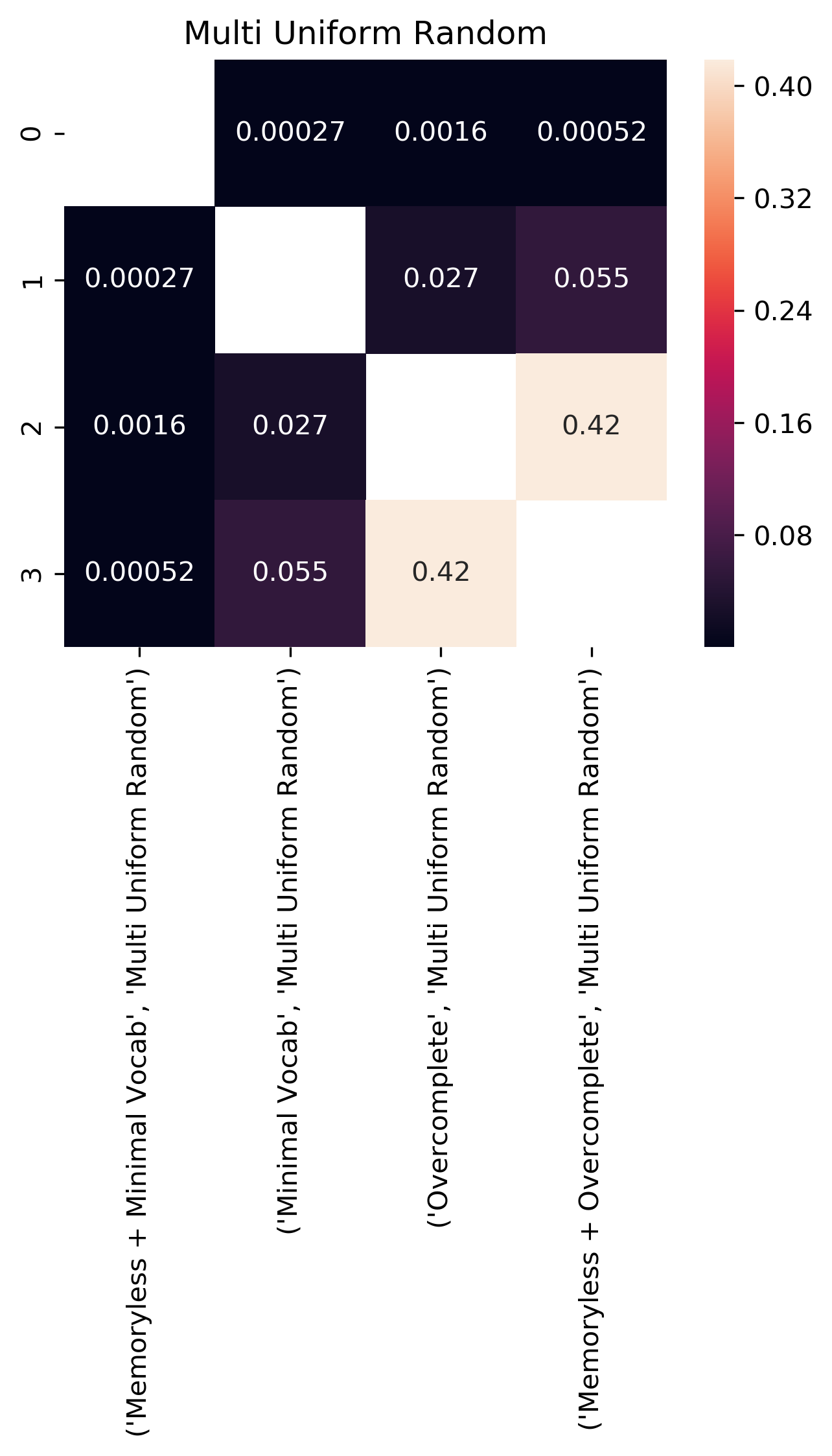}
    \end{minipage} \\
    \begin{minipage}{0.4\textwidth}
        \centering
        \includegraphics[width=\linewidth]{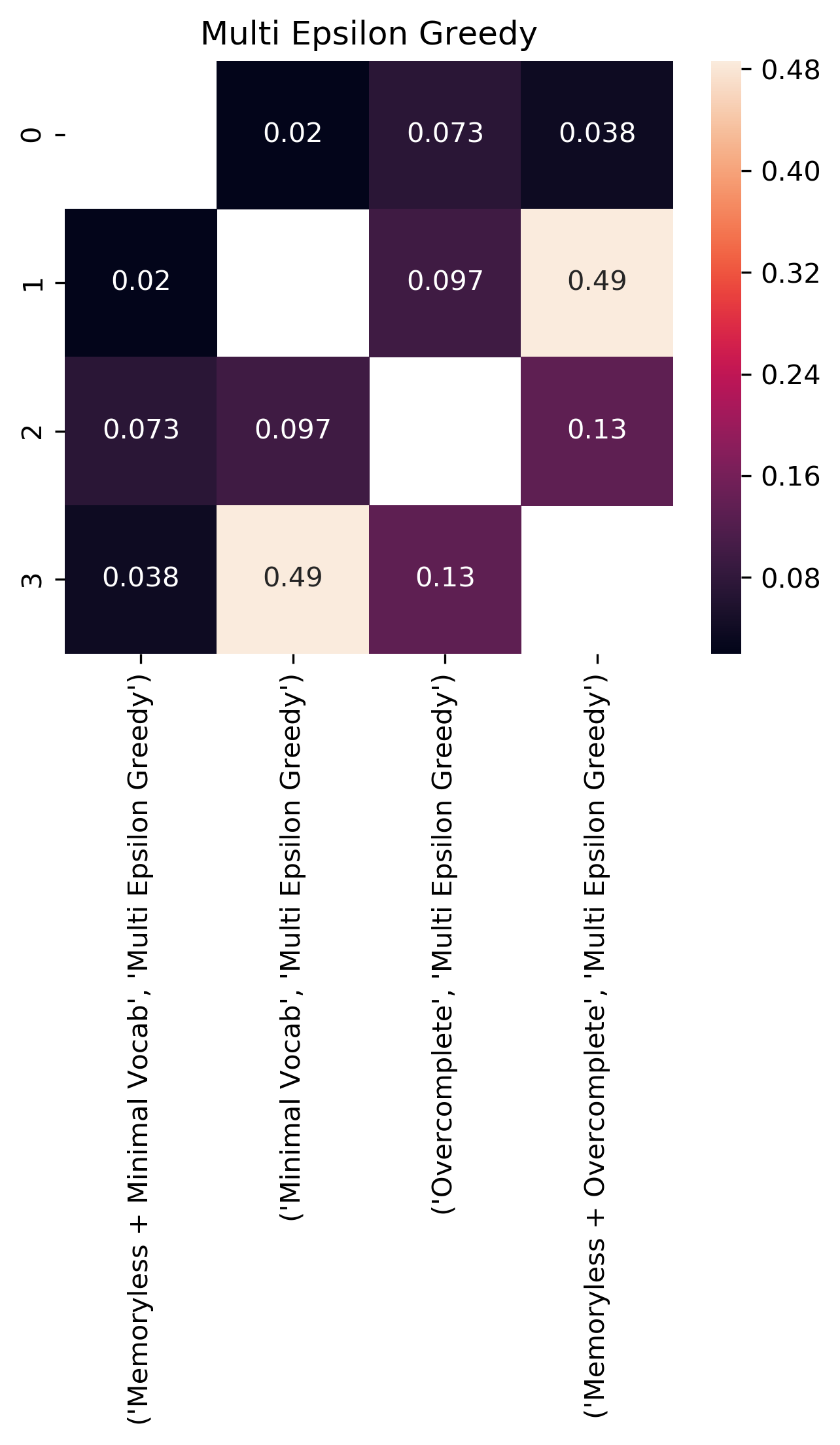}
    \end{minipage}
    \begin{minipage}{0.4\textwidth}
        \centering
        \includegraphics[width=\linewidth]{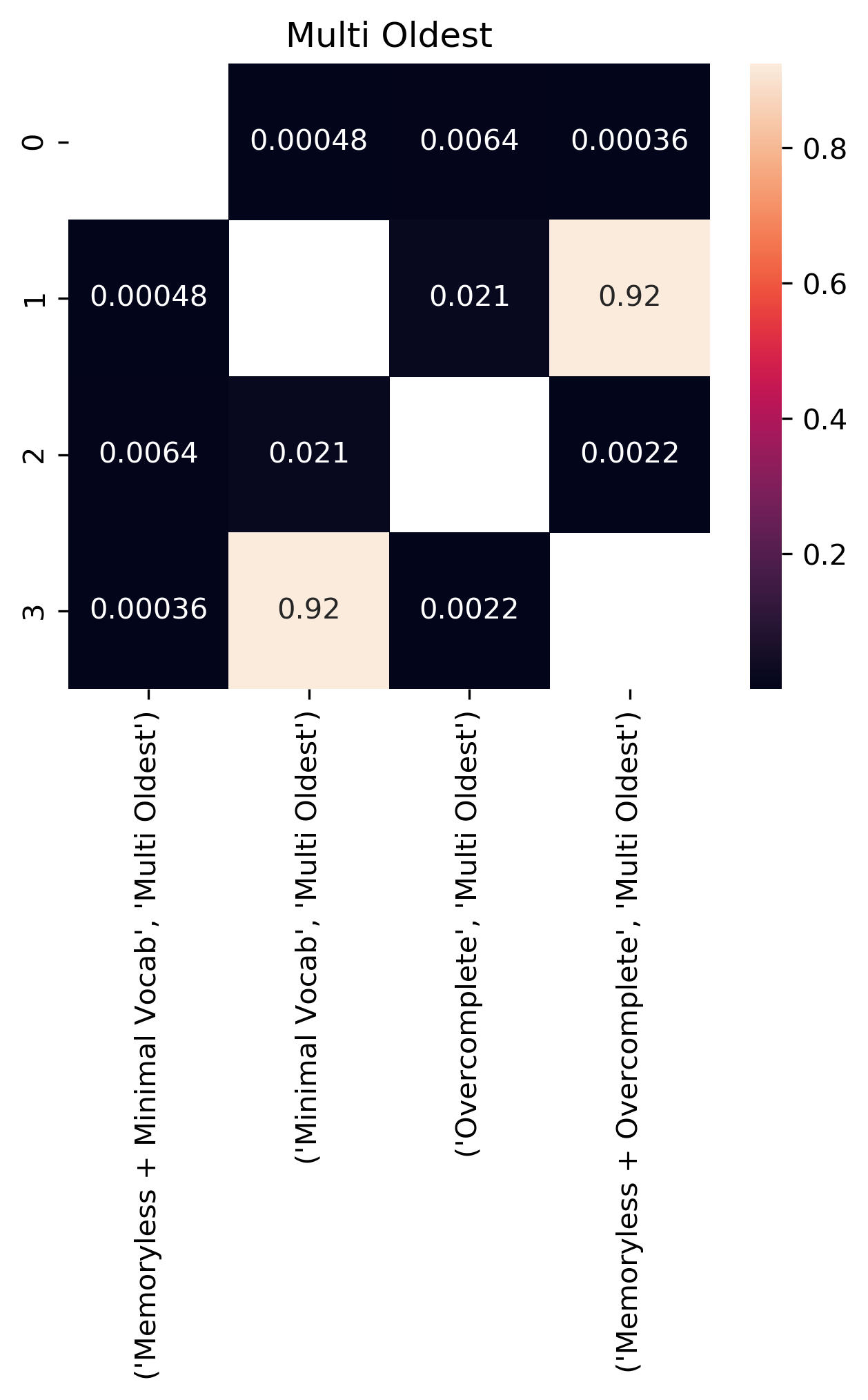}
    \end{minipage}
    \caption{Multi Agent model comparison p-values.}
    \label{fig:pvalue_method_multi}
\end{figure*}

Replacement strategy comparisons are in
\figref{fig:pvalue_method_single} (Single Agent) and
\figref{fig:pvalue_method_multi} (Multi Agent).
Model comparisons are in
\figref{fig:pvalue_model}.

\end{document}